\pdfoutput=1
\documentclass{ieeeaccess}
\usepackage{cite}
\usepackage{amsmath,amssymb,amsfonts}
\usepackage{algorithmic}
\usepackage{graphicx}
\usepackage{textcomp}

\usepackage{url}
\usepackage{multirow}
\usepackage{tabularx}

\usepackage{soul}

\usepackage[caption=false]{subfig}
\usepackage{caption,setspace}
\usepackage{textcomp}

\usepackage[norelsize]{algorithm2e}

\definecolor{newcolor}{rgb}{.8,.349,.1}

\def\BibTeX{{\rm B\kern-.05em{\sc i\kern-.025em b}\kern-.08em
    T\kern-.1667em\lower.7ex\hbox{E}\kern-.125emX}}
\begin{document}
\history{Date of publication xxxx 00, 0000, date of current version xxxx 00, 0000.}
\doi{10.1109/ACCESS.2017.DOI}

\title{GET-AID: Visual Recognition of Human Rights Abuses via Global Emotional Traits}

\author{\uppercase{GRIGORIOS KALLIATAKIS}\authorrefmark{1}, \uppercase{SHOAIB EHSAN}\authorrefmark{1}, \uppercase{MARIA FASLI\authorrefmark{1}, and KLAUS D. MCDONALD-MAIER}.\authorrefmark{1},
\IEEEmembership{Senior Member, IEEE)}}
\address[1]{School of Computer Science and Electronic Engineering, University of Essex, Colchester CO4 3SQ, U.K. }

\tfootnote{This work was supported by the Economic and Social Research Council (ESRC) under grant ES/ M010236/1, and Engineering and Physical Sciences Research Council
(EPSRC) under grants EP/R02572X/1 and EP/P017487/1.}

\markboth
{Author \headeretal: Preparation of Papers for IEEE TRANSACTIONS and JOURNALS}
{Author \headeretal: Preparation of Papers for IEEE TRANSACTIONS and JOURNALS}

\corresp{Corresponding author: Grigorios Kalliatakis (e-mail: gkallia@essex.ac.uk).}

\begin{abstract}
In the era of social media and big data, the use of visual evidence to document conflict and human rights abuse has become an important element for human rights organizations and advocates. In this paper, we address the task of detecting two types of human rights abuses in challenging, everyday photos: (1) child labour, and (2) displaced populations. We propose a novel model that is driven by a human-centric approach. Our hypothesis is that the emotional state of a person {\textendash} how positive or pleasant an emotion is, and the control level of the situation by the person {\textendash} are powerful cues for perceiving potential human rights violations. To exploit these cues, our model learns to predict global emotional traits over a given image based on the joint analysis of every detected person and the whole scene. By integrating these predictions with a data-driven convolutional neural network (CNN) classifier, our system efficiently infers potential human rights abuses in a clean, end-to-end system we call \textit{GET-AID} (from Global Emotional Traits for Abuse IDentification). Extensive experiments are performed to verify our method on the recently introduced subset of Human Rights Archive (HRA) dataset (2 violation categories with the same number of positive and negative samples), where we show quantitatively compelling results. Compared with previous works and the sole use of a CNN classifier, this paper improves the coverage up to 23.73\% for child labour and 57.21\% for displaced populations. Our dataset, codes and trained models are available online at \ul{https://github.com/GKalliatakis/GET-AID}.
\end{abstract}


\titlepgskip=-15pt

\maketitle

\section{Introduction}
\label{sec:introduction}
In the era of mobile phones with photo and video capabilities, social media and the Internet, citizen media and other publicly available footage can provide documentation of human rights violations and war crimes. However, the omnipresence of visual evidence may deluge those accountable for analysing it. Currently, information extraction from human-rights-related video and imagery requires manual labour by human rights analysts and advocates. Such analysis is utterly expensive and time consuming, while it is emotionally traumatic to focus solely on images of horrific events without context.
This paper attempts to address the problem of predicting human rights abuses from a single image, and provides technical insight and solutions to address the above issues. Note that naive schemes based on object detection or scene recognition are doomed to fail in this binary classification problem as illustrated in Fig. \ref{Fig. 1}. 
If we can exploit existing smartphone cameras, which are ubiquitous, it may be possible to turn human rights image analysis into a powerful and cost-effective computer vision application.

Recent work by Kalliatakis \textit{et al.} \cite{kalliatakis2018exploring} has shown that fine-tuned deep CNNs can address the human rights abuses classification problem to a certain extent. In this paper, we investigate the potency of representation learning methods in solving two independent, binary classification problems in the context of human rights image analysis:

\begin{enumerate}
	\item child labour / no child labour
	\item displaced populations / no displaced populations
\end{enumerate}

As will be shown in the experimental section, we found that the integration of novel \textit{global emotional traits} with a fine-tuned CNN reports an improvement in coverage up to 23.73\% for child labour and 57.21\% for displaced populations, over the sole use of a CNN classifier that is trained end-to-end using the given training data. Based on these new global emotional traits, our approach makes the following technical contributions:

\begin{figure}
	\centering
	\begin{tabular}{cc}
		\includegraphics[width=0.22\textwidth,height=0.43\textheight,keepaspectratio]{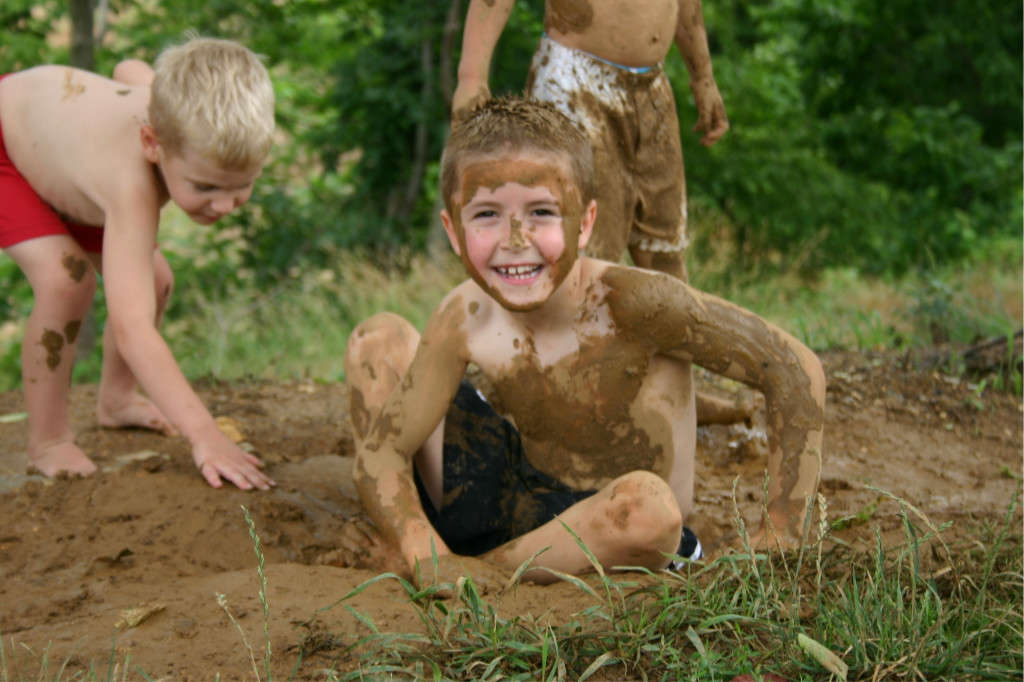} &   \includegraphics[width=0.22\textwidth,height=0.43\textheight,keepaspectratio]{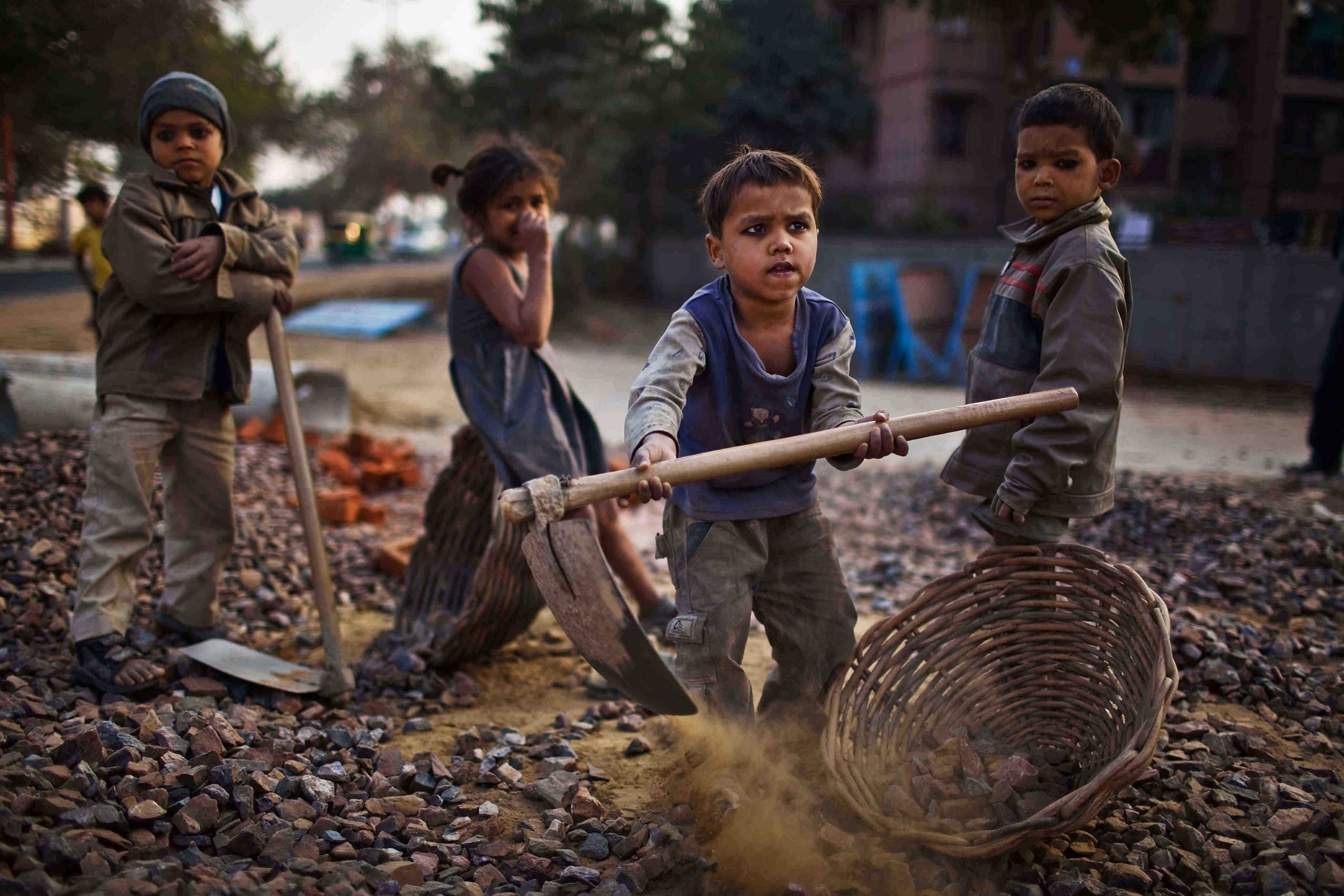} \\
		(a) Children playing. & (b) Child labour. \\[7pt]
		\includegraphics[width=0.22\textwidth,height=0.43\textheight,keepaspectratio]{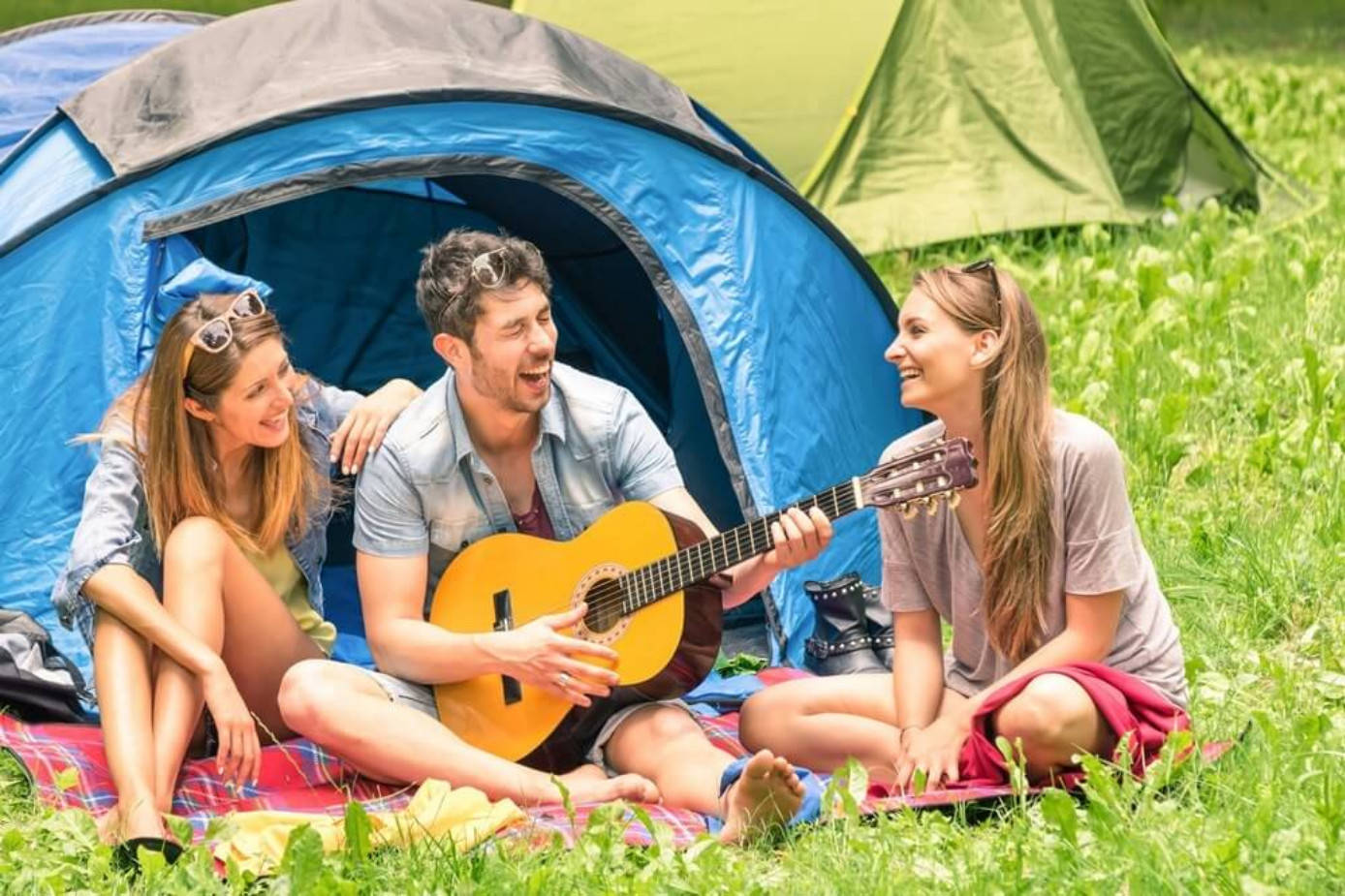} &   \includegraphics[width=0.22\textwidth,height=0.43\textheight,keepaspectratio]{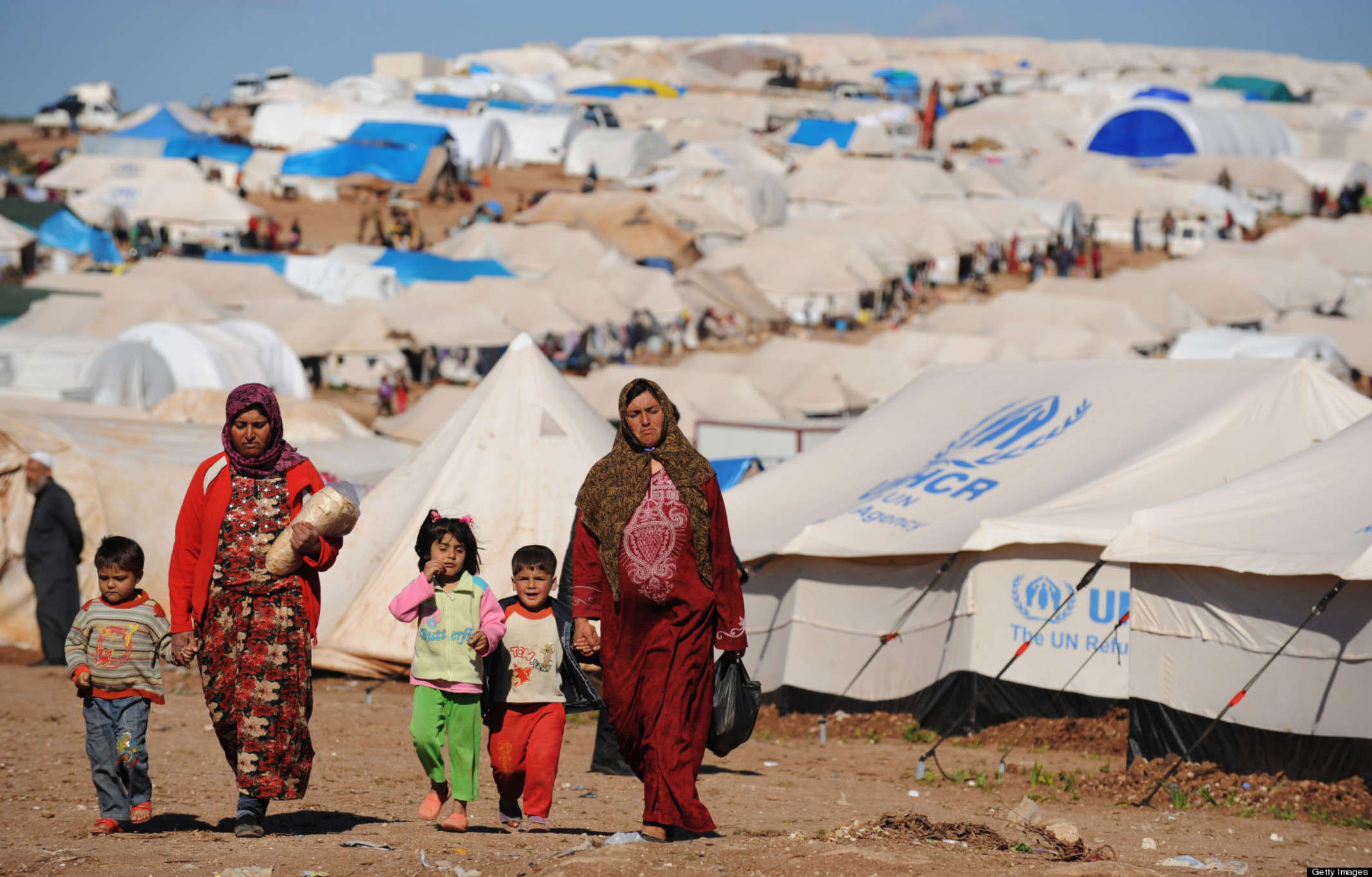} \\
		(c) Camping. & (d) Displaced populations. \\[7pt]
	\end{tabular}
	\caption{In many instances, emotional traits can be the only notifying difference between an image that depicts human rights abuse and a non-violent situation.}
	\label{Fig. 1}
\end{figure}

First, we describe the design and implementation of an end-to-end system, which is used to estimate the emotional states of people, adopting the continuous dimensions of the \textit{VAD Emotional State Model} \cite{mehrabian1995framework}. This model describes emotions using three numerical dimensions: Valence (V) to measure how positive or pleasant an emotion is; Arousal (A) to measure the agitation level of the person; and Dominance (D) that measures the control level of the situation by the person.

Our second technical contribution comprises a new mechanism capable of characterising an image based on the emotional states of all people in the scene, termed \textit{global emotional traits (GET)}. This mechanism exploits two of the continuous dimensions of the VAD emotional state model which are relevant to human rights image analysis. As will be explained in the following, global emotional traits are learned by jointly analysing each person and the entire scene similar to \cite{kosti2017emotion}.

We present a human-centric, end-to-end model for recognising two types of human rights violations. Our central observation is that the proposed global emotional traits of an image, which expose the entire mood of a situation are highly informative for inferring potential human rights abuses, when used along with a standard image classification system. Finally, we describe \textit{HRA\textemdash Binary}, a subset of the HRA dataset \cite{kalliatakis2018exploring} as benchmark for this classification task. This is used to evaluate our GET-AID model.

The remainder of the paper is organized as follows. In Section \ref{sec:related_work} we review the related work on object detection, emotion recognition and human rights abuses recognition. In Section \ref{sec:method}, we introduce our method, GET-AID. 
In Section \ref{sec:dataset_and_metrics}, we describe the subset of the HRA dataset. Section \ref{sec:experiments} discusses our results on human rights abuse classification. We conclude this paper in Section \ref{sec:conclusion}.

\section{Related work}
\label{sec:related_work}
This section gives an overview of the related work on human rights abuses recognition, which can be regarded as a category in image understanding. The background of object detection and emotion recognition are also investigated since they are important components in our human-centric approach.

\subsection{Human rights abuses understanding}
\label{subsec:human_rights_abuses_understanding}

Human rights abuses understanding plays a crucial role in human rights advocacy and accountability efforts. Automatic perception of potential human rights violations enables to discover content, that may otherwise be concealed by sheer volume of visual data. These automated systems are not producing evidence, but are instead narrowing down the volume of material that must be examined by human analysts who are making legitimate claims, that they then present in justice, accountability, or advocacy settings \cite{aronson2018computer}.

One recent example of the potential use of AI to advance human rights analysis can be found in the work of the Center for Human Rights Science (CHRS) at Carnegie Mellon University  \cite{piraces_2018}, where they have created computer vision methods to rapidly process and analyse large amount of video by detecting audio (explosions and gunshots), detecting and counting people and synchronising multiple videos from different sources. Another example is the Event Labeling through Analytic Media Processing (E-LAMP) computer vision-based video analysis system \cite{aronson2015video} which is capable of detecting objects, sounds, speech, text, and event types in a video collection. 

Though the above methods have shown good performance in their respective applications, a significant concern associated with human rights-related technology, is that the ability to extract large video collections tends to be limited to institutions with large staff base or access to expensive, technologically advanced tools and techniques \cite{Kalliatakis2017APS}. A different group of methods based on a single given image alongside the first ever image dataset for the purpose of human rights violations recognition was introduced in \cite{visapp17}. Recently, Kalliatakis \textit{et al.} \cite{kalliatakis2018exploring} proposed to recognise human rights violations from a single image by fine-tuning object-centric and scene-centric CNNs on a larger, verified-by-experts image dataset. That work also introduced a web-demo for recognising human rights violations, accessible through computer or mobile device browsers.

\begin{figure*}[t!]
	\centering
	\includegraphics[width=0.90\textwidth,height=0.29\textheight,keepaspectratio]{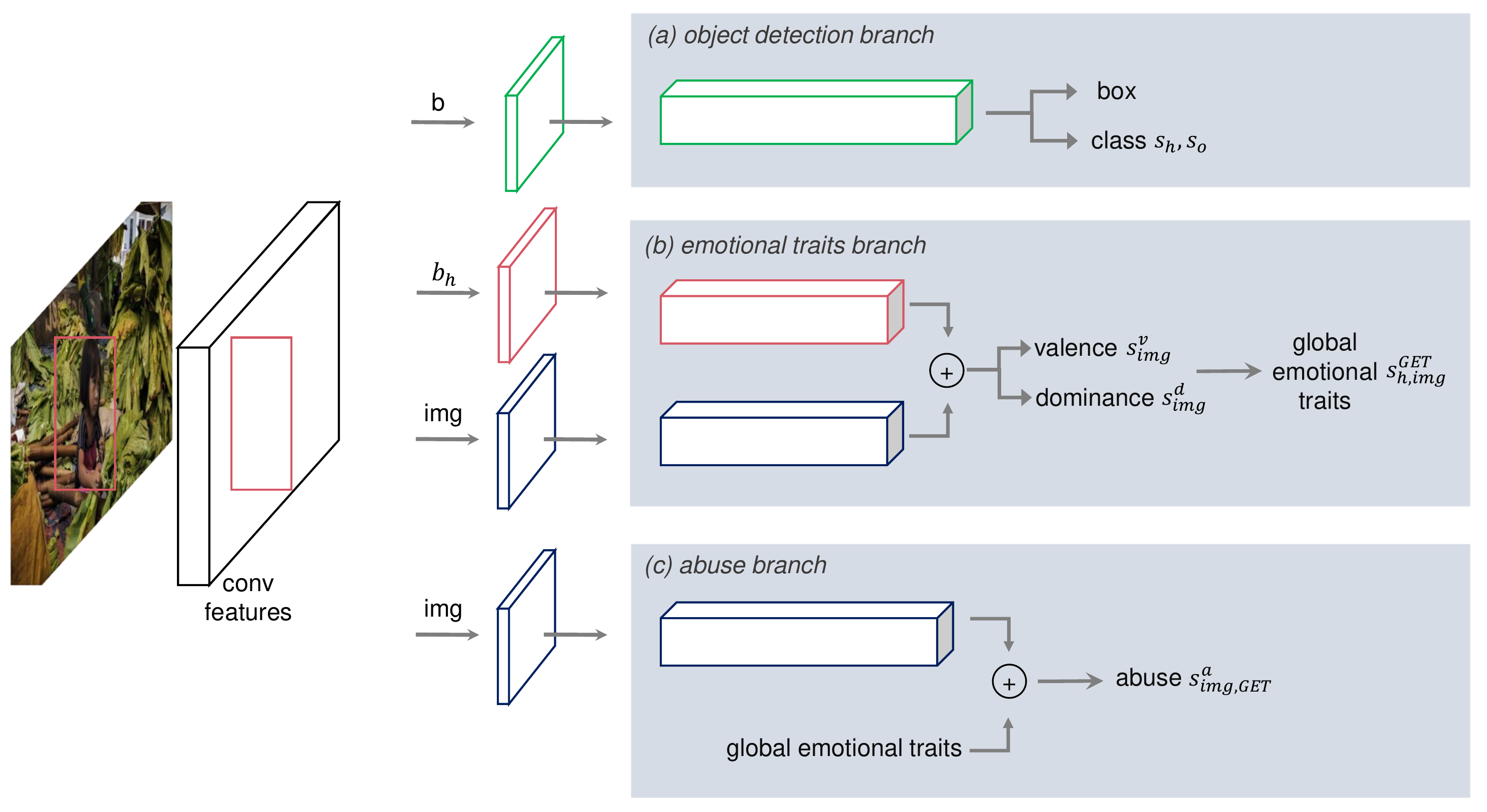}
	\caption{Model Architecture. Our model consists of (a) an object detection branch, (b) an emotional traits branch, and (c) an abuse-centric branch. The image features and their layers are shared between the emotional traits and abuse branches (blue boxes).}
	\label{Fig. 2}
\end{figure*}

\subsection{Object detection}
\label{subsec:object_detection}
Object detection, the process of determining the instance of the class to which an object belongs and estimating the location of the object by outputting the bounding box around the object, is one of the computer vision areas that has improved steadily in the past few years. It is possible to group object detectors using different aspects, but one of the most accepted categorisation is to split object detectors into two categories: one-stage detectors and two-stage detectors.

\textbf{One-stage detectors:} The OverFeat model \cite{OverFeat} which applies a sliding window approach based on multi-scaling for jointly performing classification, detection and localization was one of the first modern one-stage object detectors based on deep networks. More recently YOLO\cite{liu2016ssd,Fu2017DSSDD} and SSD \cite{redmon2016you,redmon2016yolo9000} have revived interest in one-stage methods, mainly because of their real time capabilities, although their accuracy trails that of two-stage methods. One of the main reasons being due to the class imbalance problem \cite{lin2018focal}.

\textbf{Two-stage detectors:} the leading model in modern object detection is based on a two-stage approach which was established in \cite{uijlings2013selective}. The first stage generates a sparse set of candidate proposals that should contain all objects, and the second stage classifies the proposals into foreground classes or background. R-CNN, a notably successful family of methods \cite{girshick2015fast, girshick2014rich} enhanced the second-stage classifier to a convolutional network, resulting in large accuracy improvements. The speed of R-CNN has improved over the years by integrating region proposal networks (RPN) with the second-stage classifier into a single convolution network, known as the Faster R-CNN framework  \cite{ren2015faster}. 

In this work, we adopt the one-stage approach of RetinaNet framework \cite{lin2018focal} which handles class imbalance by reshaping the standard cross entropy loss to focus training on a sparse set of hard examples and down-weights the loss assigned to well-classified examples.

\begin{figure*}[t!]
	\centering
	\begin{tabular}{cc}
		\includegraphics[width=0.56\textwidth,height=0.25\textheight,keepaspectratio]{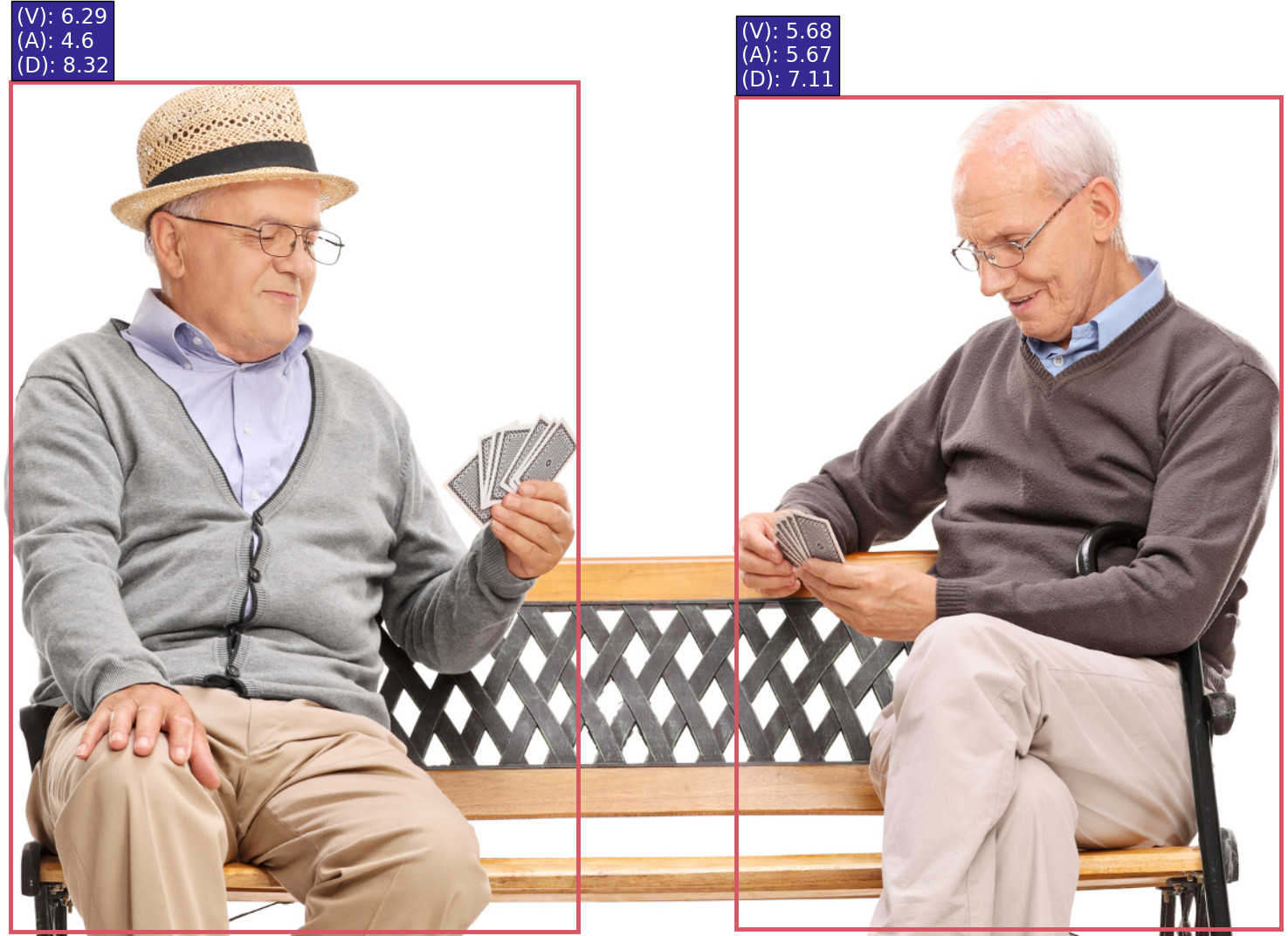} &   \includegraphics[width=0.56\textwidth,height=0.25\textheight,keepaspectratio]{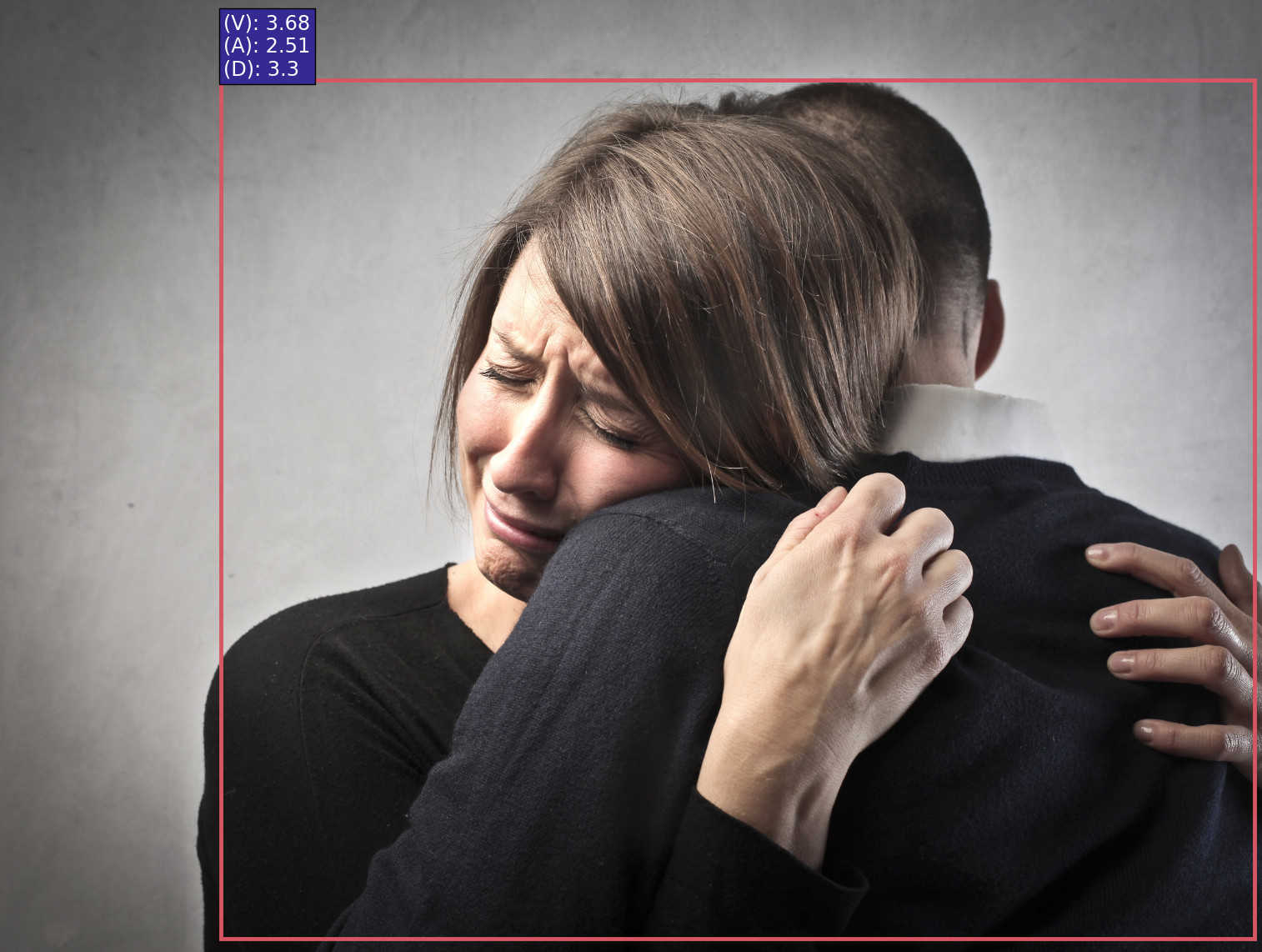} \\
		(a) & (b)  \\[6pt]
	\end{tabular}
	\caption{Examples of people marked with the red bounding box that have been labelled with different scores of Valence, Arousal and Dominance.}
	\label{Fig. 3}
\end{figure*}

\subsection{Emotion recognition}
\label{subsec:emotion_recognition}
Recognising people's emotional states from images is an active research area among the computer vision community. One established categorisation of emotion recognition methods in the literature is to split them into two categories based on how emotions are represented: discrete categories and continuous dimensions. 

\textbf{Discrete categories:} most of the research in computer vision to recognise people's emotional states is explored by facial expression analysis \cite{fabian2016emotionet,eleftheriadis2016joint}, where a large variety of methods have been developed to recognise the 6 basic emotions defined in \cite{ekman1971constants}. Many of these methods are based on a set of specific localised movements of the face, called \textit{Action Units}, in order to encode the facial expressions \cite{friesen1978facial,chu2017selective}. More recently emotion recognition systems based on facial expressions use CNNs to recognise the Action Units \cite{fabian2016emotionet}.

\textbf{Continuous dimensions:} instead of recognising discrete emotion categories, this family of methods use the continuous dimensions of the VAD \textit{Emotional State Model} \cite{mehrabian1995framework,mehrabian1974approach} to represent emotions. The VAD model uses a 3-dimensional approach to describe and measure the emotional experience of humans: Valence (V) describes affective states from highly negative (unpleasant) to highly positive (pleasant); Arousal (A) measures the intensity of affective states ranging from calm to excited or alert; and Dominance (D) represents the feeling of being controlled or influenced by external stimuli. In recent times the VAD model has been utilised for facial expression recognition \cite{soleymani2016analysis}.

In this paper, we adopted the tridimensional model of affective experience alongside a joint analysis of the person and the entire scene in order to recognise rich information about emotional states, similar to \cite{kosti2017emotion}. 

\section{Method}
\label{sec:method}

We now describe our method for detecting two types of human rights abuses based on the global emotional traits of people within the image. Our goal is to label challenging everyday photos as either human-rights-abuse positive (`child labour' or `displaced populations') or human-rights-abuse negative (`no child labour' or `no displaced populations' respectively). 

To detect the global emotional traits of an image, we need to accurately localise the box containing a $human$ and the associated object of interaction (denoted by $b_h$ and $b_o$, respectively), as well as identify the emotional states $e$ of each human using the VAD model. Our proposed solution adopts the RetinaNet \cite{lin2018focal} object detection framework alongside an additional \textit{emotional traits} branch that estimates the continuous dimensions of each detected person and then determines the global emotional traits of the given image. 

Specifically, given a set of candidate boxes, RetinaNet outputs a set of object boxes and a class label for each box. While the object detector can predict multiple class labels, our model is concerned only with the `person' class. The region of the image comprising the person whose feelings are to be estimated at $b_h$ is used alongside the entire image for simultaneously extracting their most relevant features. These features, are fused and used to perform continuous emotion recognition in VAD space. Our model extends typical image classification by assigning a triplet score $S_{img,GET}^a$  to pairs of candidate human boxes $b_h$ and an abuse category \textit{a}. To do so, we decompose the triplet score into three terms:

\begin{equation} \label{eq:1}
S_{img,GET}^a = S_h  \cdot S_{h,img}^{GET}  \cdot S_{img}^a
\end{equation}

\noindent
We discuss each component next, followed by details for training and inference. Fig. \ref{Fig. 2} illustrates each component in our full framework.

\vskip 0.2in

\subsection{Model components}
\label{subsec:model_components}

\subsubsection{Object detection}
\label{subsubsec:object_detection}
The \textit{object detection} branch of our network, shown in Fig. \ref{Fig. 2}(a), is identical to that of RetinaNet \cite{lin2018focal} single stage classifier, although the SSD \cite{liu2016ssd} detector has also been tested. First, an image is forwarded through ResNet-50 \cite{he2016deep}, then in the subsequent pyramid layers, the more semantically important features are extracted and concatenated with the original features for improved bounding box regression. Although the combination of the proposal and classification components into a single network results in a more complicated training process and lower accuracy compared to their two-stage counterparts, RetinaNet manages to counterbalance that with higher detection speed by utilising a smarter loss function named focal loss.

\begin{figure*}
	\centering
	\begin{tabular}{cc}
		\includegraphics[width=0.35\textwidth,height=0.3\textheight,keepaspectratio]{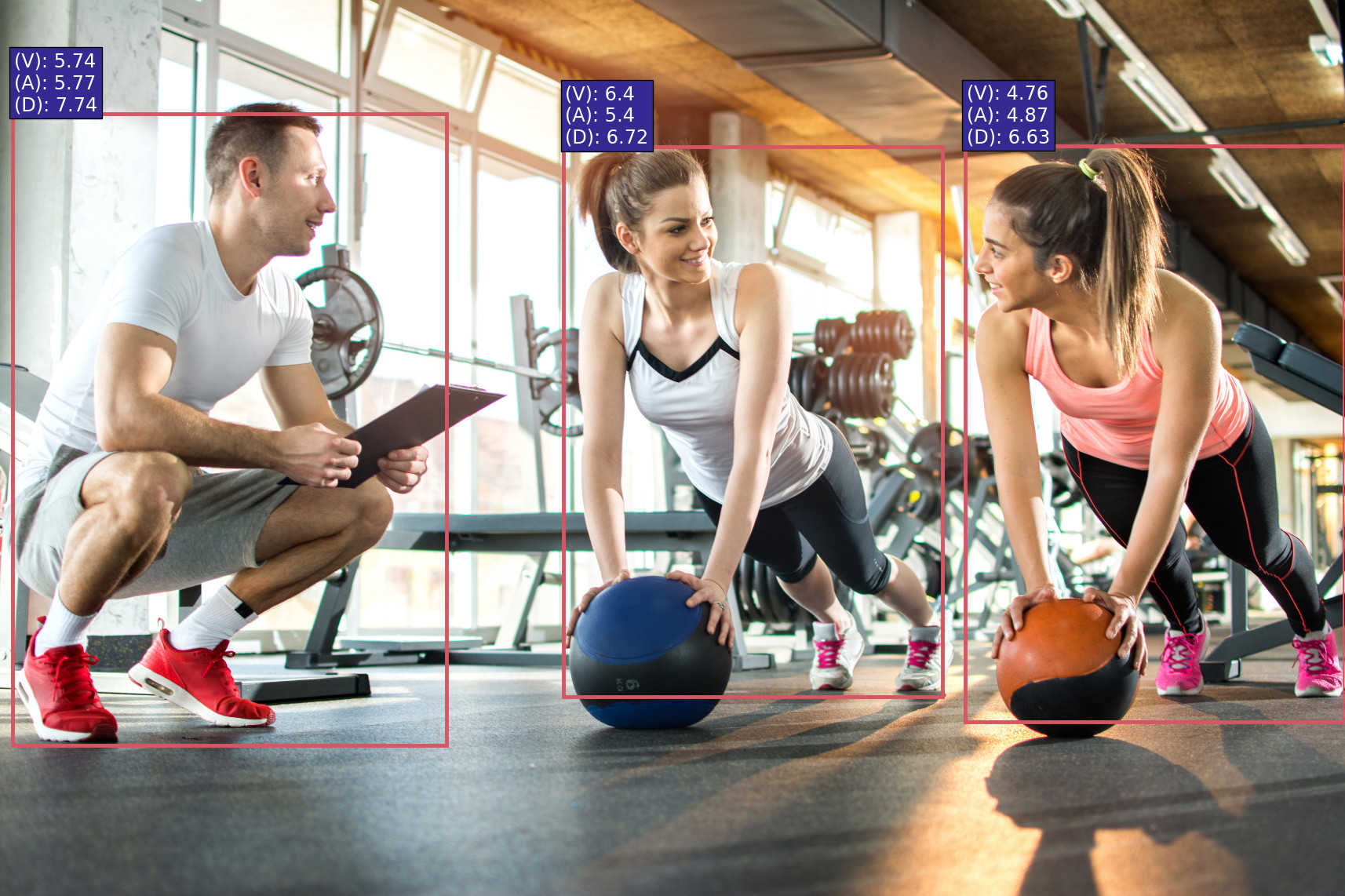} &   \includegraphics[width=0.35\textwidth,height=0.3\textheight,keepaspectratio]{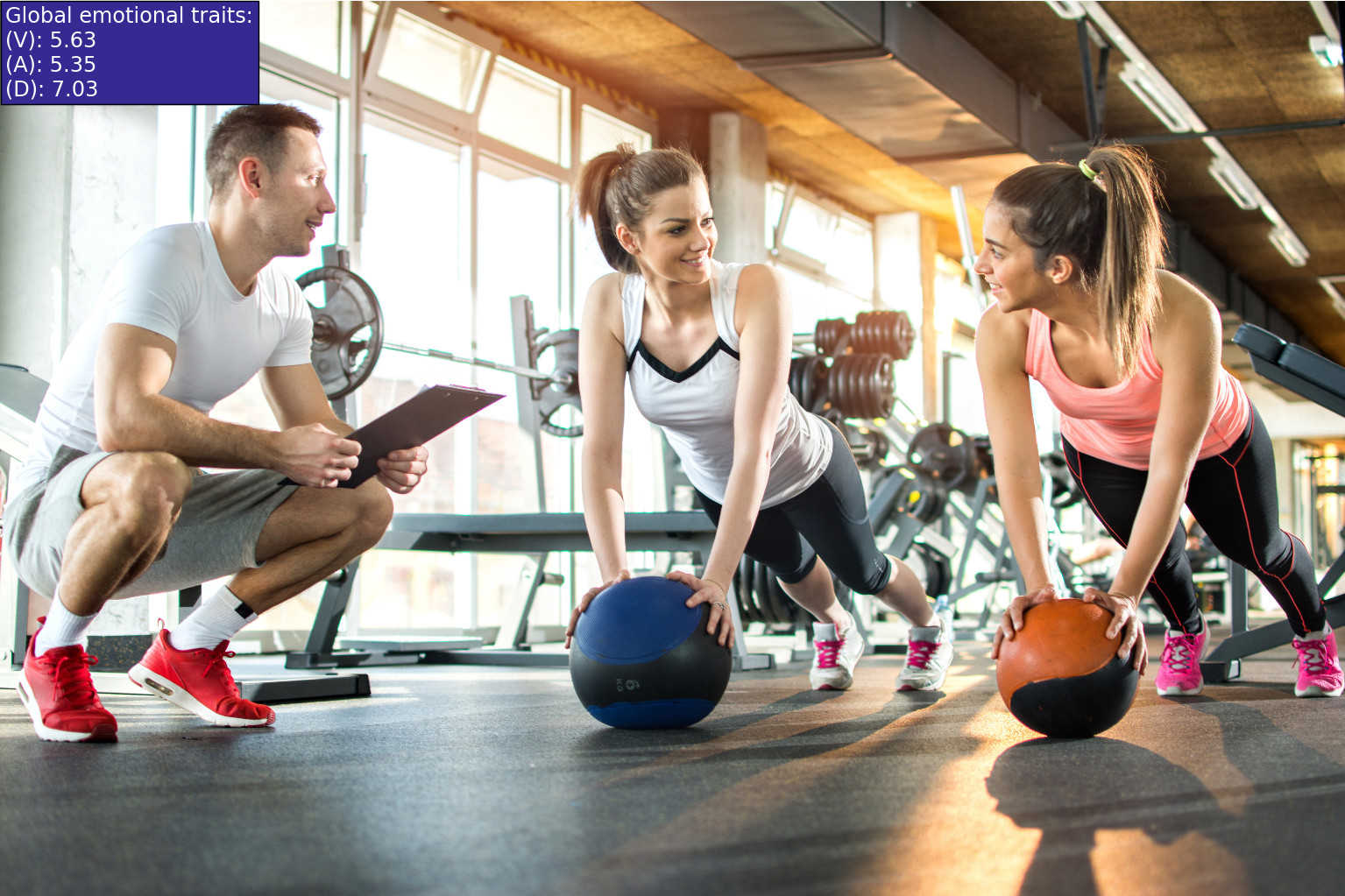} \\
		(a) & (b) \\[6pt]
		\includegraphics[width=0.35\textwidth,height=0.3\textheight,keepaspectratio]{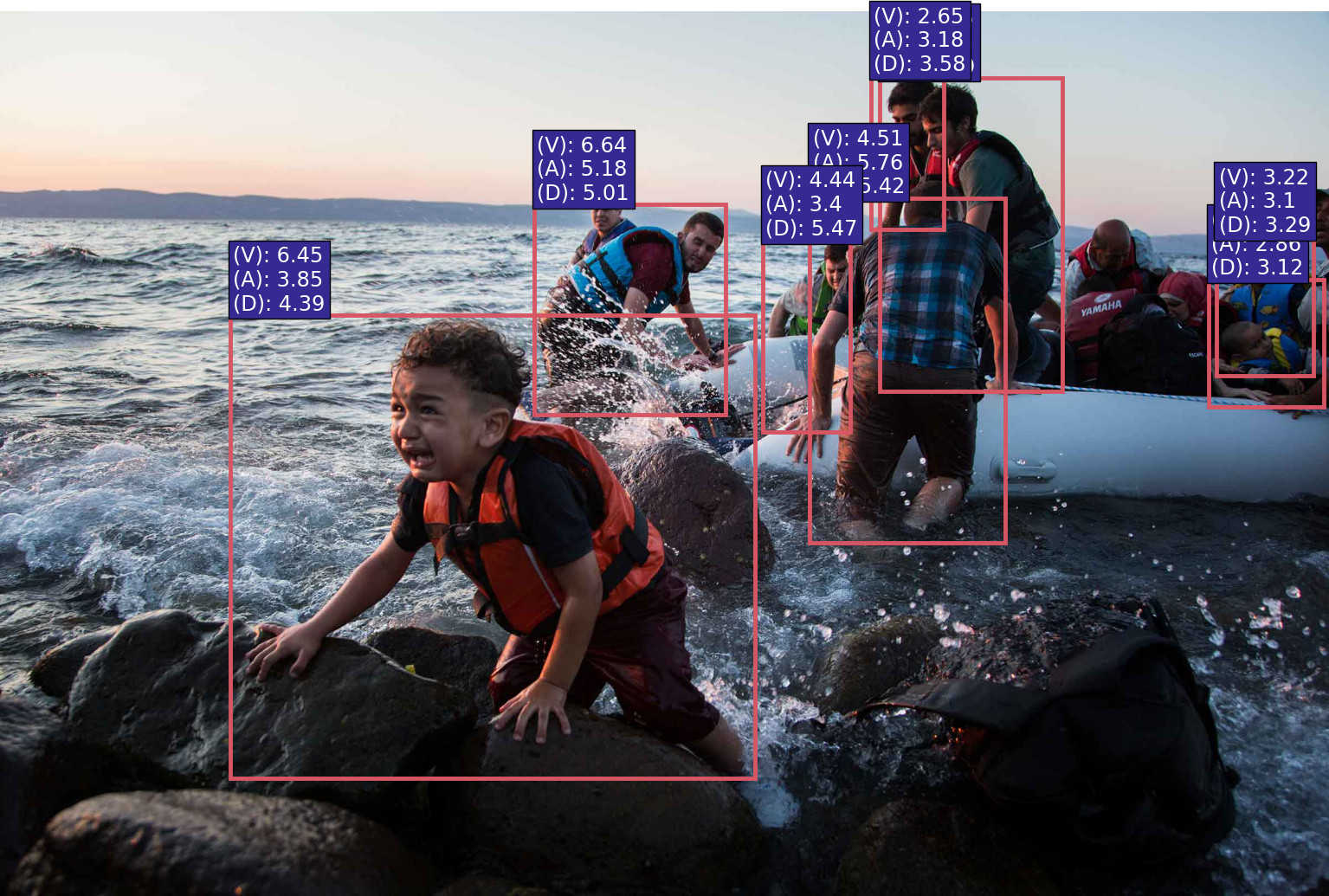} &   \includegraphics[width=0.35\textwidth,height=0.3\textheight,keepaspectratio]{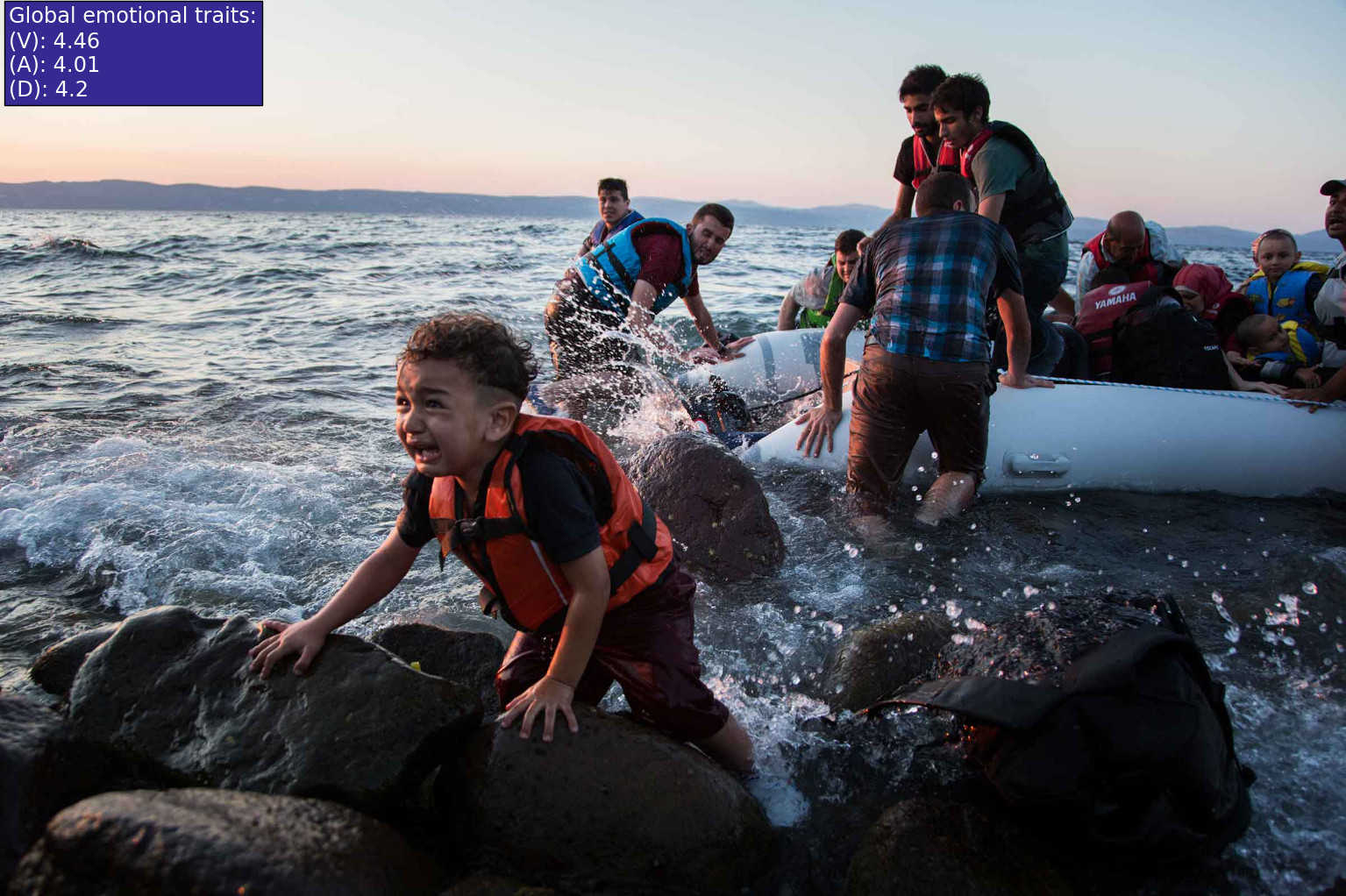} \\
		(c) & (d) \\[6pt]
	\end{tabular}
	\caption{Estimating continuous emotions in VAD space v global emotional traits from the combined body and image features. These global emotional traits will be integrated with the standard image classification scores $S_{img}^a$ to detect human rights abuses. The left column shows the predicted emotional states and their scores from the person region of interest (RoI), while the right column show the same images analysed for global emotional traits.}
	\label{Fig. 4}
\end{figure*}

\subsubsection{Emotional traits recognition}
\label{subsubsec:emotional_traits_recognition}
The first role of the \textit{emotional traits} branch is to assign an emotion classification score $S_{h,img}^e$ to each human box $b_h$ and emotion $e$. Similar to \cite{kosti2017emotion}, we use an end-to-end model with three main modules: two feature extractors and a fusion module. The first module takes the region of the image comprising the person whose emotional traits are to be estimated, $b_h$, while the second module takes as input the entire image and extracts global features. This way the required contextual support is accommodated in our emotion recognition process. Finally, the third module takes as input the extracted image and body features and estimates the continuous dimensions in VAD space. Fig. \ref{Fig. 3} shows qualitative results of the method.

The second role of the \textit{emotional traits} branch is to assign a two-dimensional emotion classification score $S_{h,img}^{GET}$ which characterises the entire input image, based on two emotion classification scores $S_{h,img}^v$  and $S_{h,img}^d$ for each human. To the best of our knowledge, this particular score, called \textit{global emotional traits (GET)}, is the first attempt to establish a method for summarising the overall mood of a situation depicted in a single image. We decompose the two-dimensional GET score into two terms:

\begin{equation} \label{eq:2}
S_{h,img}^{GET} = S_{img}^{v}  \cdot S_{img}^{d}
\end{equation}

\noindent
In the above, $S_{img}^{v}$  is the encoding of the global valence score relative to human box $b_h$ and entire image $img$, that is:

\begin{equation} \label{eq:3}
S_{img}^{v} = \frac{1}{n} \sum_{i=1}^{n} S_{h,img}^v 
\end{equation}

\noindent
Similarly, $S_{img}^{d}$  is the encoding of the global dominance score relative to human box $b_h$ and entire image $img$, that is:

\begin{equation} \label{eq:4}
S_{img}^{d} = \frac{1}{n} \sum_{i=1}^{n} S_{h,img}^d  
\end{equation}

In Fig. \ref{Fig. 4} (a),(c) the three different emotional states over target objects location are estimated, while (b), (d) illustrate the global emotional traits proposed here. For the sake of completeness all three predicted numerical dimensions are depicted. However, only valence and dominance are considered to be relevant to human rights violations recognition tasks since the agitation level of a person, denoted by arousal, can be ambiguous for several situations. For example, both Fig. \ref{Fig. 4} (a) and (c) depict people with arousal values close to 5.5, but the activities captured are utterly different from a human rights image analysis perspective.

\subsubsection{Abuse recognition}
\label{subsubsec:abuse_recognition}
The first role of the abuse branch, shown in Fig. \ref{Fig. 2}(c), is to assign an abuse  classification score to the input image. Just like in the two-phase
transfer learning scheme deployed previously for HRA-CNNs \cite{kalliatakis2018exploring}, we train an end-to-end model for binary classification of everyday photos as either human-rights-abuse positive (`child labour' or `displaced populations') or human-rights-abuse negative (`no child labour' or `no displaced populations').

In order to improve the discriminative power of our model, the second role of the abuse branch is to integrate $S_{h,img}^{GET}$ in the recognition pipeline. Specifically, the raw image classification score $S_{img}^a$ is readjusted based on the recognised global emotional traits. Each GET unit, that is deltas from the neutral state, is expressed as a numeric weight varying between 1 and 10, while the neutral states of GET are assigned between 4.5 and 5.5 based on the number of examples per each of the scores in the continuous dimensions reported in \cite{kosti2017emotion}. The GET feature of each input image can be written in the form of a 2-element vector:

\begin{equation} \label{eq:5}
\bar{d} = (D_1, D_2) 
\end{equation}

\noindent
where $D_1$ and $D_2$ refer to the weights of valence and dominance, respectively. The adjustment that will be assigned to the raw probability, $S_{img}^a$ is the weight of valence or dominance multiplied by a factor of 0.11 which has been experimentally set. When the input image depicts positive valence or positive dominance, the adjustment factor is subtracted from the positive human-rights-abuse probability and added to the negative human-rights-abuse probability. Similarly, when the input image depicts negative valence or negative dominance the adjustment factor is added to the negative human-rights-abuse probability and subtracted from the positive human-rights-abuse probability. This is formally written in Algorithm \ref{algorithm}.

Finally, when no $b_h$ were detected from the object detection branch, (\ref{eq:1}) is reduced into plain image classification as follows:

\begin{equation} \label{eq:6}
S_{img,GET}^a = S_{img}^a 
\end{equation}

\subsection{Training}
\label{subsec:training}
We approach human rights abuses classification as a \textit{cascaded, multi-task} learning problem. Due to different datasets, convergence times and loss imbalance, all three branches have been trained separately. 
For object detection we adopted an existing implementation of the RetinaNet object detector, pre-trained on the COCO dataset \cite{lin2014microsoft}, with a ResNet-50 backbone. 

For emotion recognition in continuous dimensions, we formulate this task as a regression problem using the Euclidean loss. The two feature extraction modules described in Section \ref{subsubsec:emotional_traits_recognition}, are designed as truncated versions of various well-known convolutional neural networks and initialised using pretrained models on two large-scale image classification datasets, ImageNet \cite{krizhevsky2012imagenet} and Places \cite{zhou2018places}. The truncated version of those CNNs removes the fully connected layer and outputs features from the last convolutional layer in order to maintain the localisation of different parts of the images which is significant for the task at hand. Features extracted from these two modules (red and blue boxes in Fig. \ref{Fig. 2}(b)) are then combined by a fusion module. This module first uses a global average pooling layer to reduce the number of features from each network and then a fully connected layer, with an output of a 256-D vector,  functions as a dimensionality reduction layer for the concatenated pooled features. Finally, we include a second fully connected layer with 3 neurons representing valence, arousal and dominance. The parameters of the three modules are learned jointly using stochastic gradient descent with momentum of 0.9. The batch size is set to 54 and we use dropout with a ratio of 0.5.

For human rights abuse classification, we formulate this task as a binary classification problem. We train an end-to-end model for classifying everyday images as human-rights-abuse positive or human-rights-abuse negative, based on the context of the images, for two independent use cases, namely \textit{child labour} and \textit{displaced populations}. Following the two-phase transfer learning scheme proposed in \cite{kalliatakis2018exploring}, we fine-tune various CNN models for the two-class classification task. First, we conduct feature extraction utilising only the convolutional base of the original networks in order to end up with more generic representations as well as retaining spatial information similar to emotion recognition pipeline. The second phase consists of unfreezing some of the top layers of the convolutional base and jointly training a newly added fully connected layer and these top layers.

All the CNNs presented here were trained using the Keras Python deep learning framework \cite{chollet2015keras} over TensorFlow \cite{abadi2016tensorflow} on Nvidia GPU P100.

\subsection{Inference}
\label{subsec:inference}
\textit{Object Detection Branch}: We first detect all objects (including the person class) in the input image. We apply a threshold on boxes with scores higher than 0.5, which is set conservatively to retain most objects. This yields a set of $n$ boxes $b$ with scores $s_h$ and $s_o$. These boxes are used as input to the emotional trait branch. 

\textit{Emotional Traits Branch}: Next, we apply the emotional traits branch to all detected objects that were classified as \textit{human}. We feed each human box $b_h$ alongside the entire input image $img$ to the VAD emotion recognition model. For each $b_h$, we predict valence $s_{v}$ and arousal $s_{a}$ scores, and then compute the \textit{global emotional traits} $S_{h,img}^{GET}$ that describes the entire input image $img$.

\textit{Abuse Branch}: If no human box $b_h$ has been detected, for example when a plain beach without people or kitchen appliances was given as input image, the branch predicts the two abuse scores $S_v$ and $S_nv$ directly from the binary classifier. On the other hand, when one or more people have been detected, the branch weights the raw predictions from the binary classifier based on the computed global emotional traits of the input image according to  Algorithm \ref{algorithm}. 

\begin{algorithm} 
	\caption{Calculate $S_{img,GET}^a$} 
	\label{algorithm} 
	\begin{algorithmic} 
		\REQUIRE $b_h > 0$
		\STATE $S_{v}\gets S_{img}^v$ \COMMENT{$v$: child labour/disp. populations}
		\STATE $S_{nv}\gets S_{img}^{nv}$ \COMMENT{$nv$: no child labour/no disp. populations}
		\IF {$D_1$ $\geq$ 4.5  \AND $D_1$ $\leq$ 5.5}
		\STATE $S_{v} = S_{img}^v$
		\STATE $S_{nv} = S_{img}^{nv}$
		\ELSIF {$D_1> 5.5$} 
		\STATE $diff = D_1-5.5$
		\STATE $adj = diff*0.11$
		\STATE $S_{v} = S_{v}-adj$
		\STATE $S_{nv} = S_{nv}+adj$
		\ELSIF {$D_1< 4.5$} 
		\STATE $diff = 4.5-D_1$
		\STATE $adj = diff*0.11$
		\STATE $S_{v} = S_{v}+adj$
		\STATE $S_{nv} = S_{nv}-adj$
		\ENDIF
		\IF {$D_2$ $\geq$ 4.5  \AND $D_2$ $\leq$ 5.5} 
		\STATE Return $S_{v}, S_{nv}$
		\ELSIF {$D_2> 5.5$} 
		\STATE $diff = D_2-5.5$
		\STATE $adj = diff*0.11$
		\STATE $S_{v} = S_{v}-adj$
		\STATE $S_{nv} = S_{nv}+adj$
		\ELSIF {$D_2< 4.5$} 
		\STATE $diff = 4.5-2_1$
		\STATE $adj = diff*0.11$
		\STATE $S_{v} = S_{v}+adj$
		\STATE $S_{nv} = S_{nv}-adj$
		\ENDIF
		\STATE Return $S_{v}, S_{nv}$
	\end{algorithmic}
\end{algorithm}

\section{Dataset and metrics}
\label{sec:dataset_and_metrics}
There are a limited number of image datasets for human rights violations recognition \cite{kalliatakis2018exploring, visapp17}. The most relevant for this work is HRA (\textit{Human Rights Archive}) \cite{kalliatakis2018exploring}. In order to find the main test platform on which we could demonstrate the effectiveness of GET-AID and analyse its various components, we construct a new image dataset by maintaining the verified samples intact for the two categories with more images, \textit{child labour} and \textit{displaced populations}. The \textit{HRA\textemdash Binary} dataset contains 1554 images of human rights abuses and the same number of no violation counterparts for training, as well as 200 images collected from the web for testing and validation. Note that each abuse category is treated as an independent use case for our experiments. The dataset is made publicly available for future research.

Following \cite{kalliatakis2018exploring}, we evaluate GET-AID with two metrics \textit{accuracy} and \textit{coverage} and compare its performance against the sole use of a CNN classifier trained on HRA\textemdash Binary. We also evaluate the continuous dimensions using error rates - the difference (in average) between the true value and the regressed value.

\vskip 0.22in

\section{Experiments}
\label{sec:experiments}

\subsection{Implementation details}
\label{subsec:implementation_details}
Our emotion recognition implementation is based on the emotion recognition in context (EMOTIC) model \cite{kosti2017emotion}, with the difference that our model estimates only continuous dimensions in VAD space. We train the three main modules (Section \ref{subsec:training}) on the EMOTIC database, which contains a total number of 18,316 images with 23,788 annotated people, using pre-trained CNN feature extraction modules. We treat this multiclass-multilabel problem as a regression problem by using a weighted Euclidean loss to compensate for the class imbalance of EMOTIC. 

Table \ref{tab1} shows the results for the continuous dimensions using error rates. The best result is obtained by utilising \textit{model ensembling}, which consists of pooling together the predictions of a set of different models in order to produce better predictions. We pool the predictions of classifiers (\textit{ensemble the classifiers}) by conducting weighted average of their prediction at inference time. The weights are learned on the validation data - usually the better single classifiers are given a higher weight, while the worse single classifiers are given a lower weight. However, ensembling the classifiers results in prolonged inference times, which causes us to turn our focus onto single classifiers for the remainder of the experiments.

\begin{table}[t!]
	\caption{Mean error rate obtained (average of all three VAD dimensions) for different body feature backbone CNNs. The image feature backbone CNN was kept constant for all cases, namely VGG16-Places365 \cite{zhou2018places}. }
	\label{tab1}
	\begin{tabular}{cc}
		\hline
		\textbf{Body Feature Backbone} & \textbf{Mean error rate} \\ \hline
		VGG16                          & 1.59                     \\
		VGG19                          & 1.57                     \\
		ResNet50                       & 1.69                     \\ \hline
		VGG16 + ResNet50               & 1.40                     \\
		VGG16 + VGG19                  & 1.36                     \\
		VGG19 + ResNet50               & 1.48                     \\ \hline
		VGG19 + ResNet50 + VGG16       & 1.36                     \\ \hline
	\end{tabular}
\end{table}

Following the two-phase transfer learning scheme proposed in \cite{kalliatakis2018exploring}, we fine-tune our human-rights-abuse classification models, Fig. \ref{Fig. 2}(c), for 50 iterations on the HRA\textemdash Binary $trainval$ set with a learning rate of 0.0001 using the stochastic gradient descent (SGD) \cite{lecun1989backpropagation} optimizer for cross-entropy minimization. These \textit{vanilla} models will be examined against GET-AID. Here, vanilla means pure image classification using solely fine-tuning without any alteration.

\begin{figure*}[t!]
	\centering
	\begin{tabular}{cccc}
		\includegraphics[width=0.24\textwidth,height=0.42\textheight,keepaspectratio]{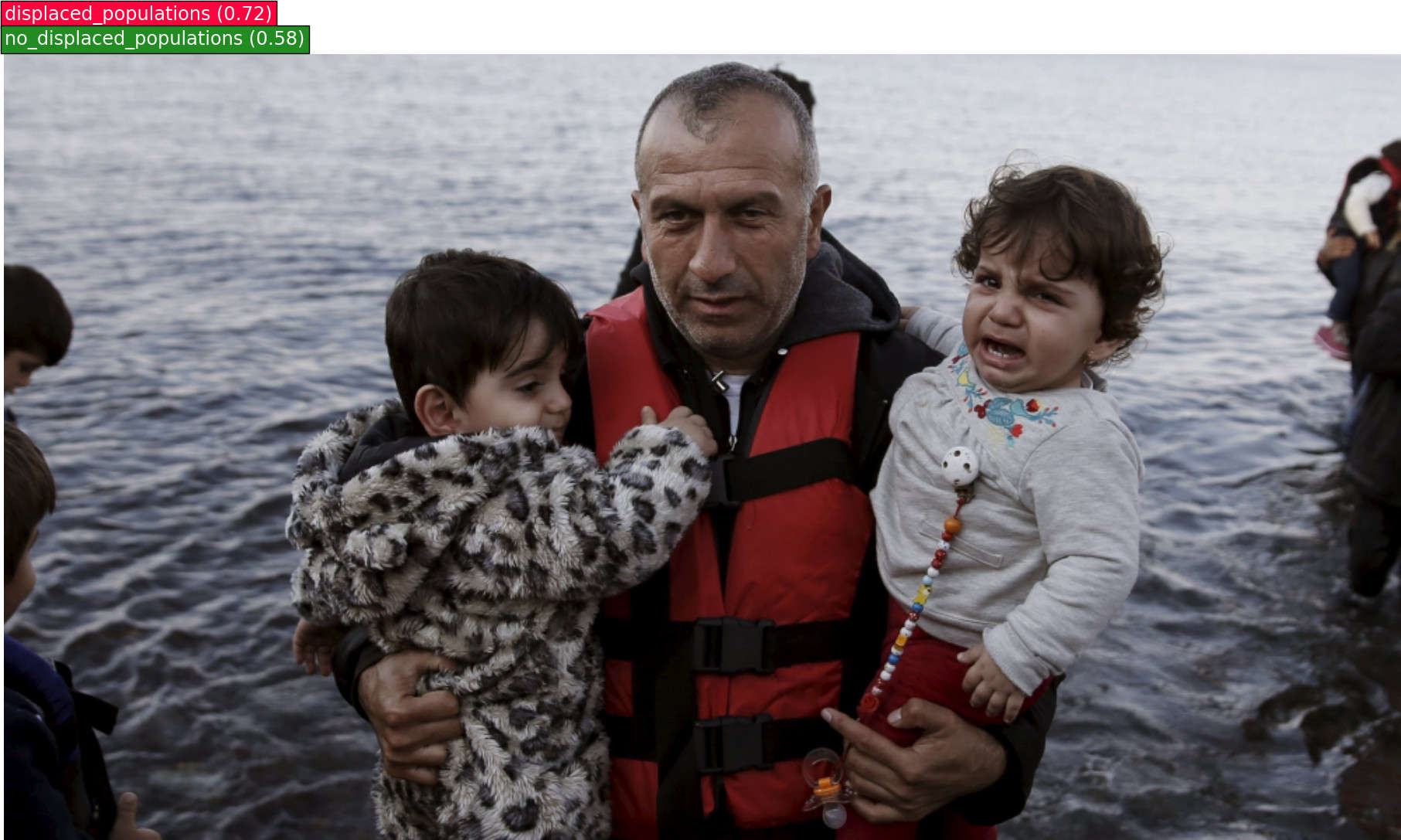} &   \includegraphics[width=0.24\textwidth,height=0.3\textheight,keepaspectratio]{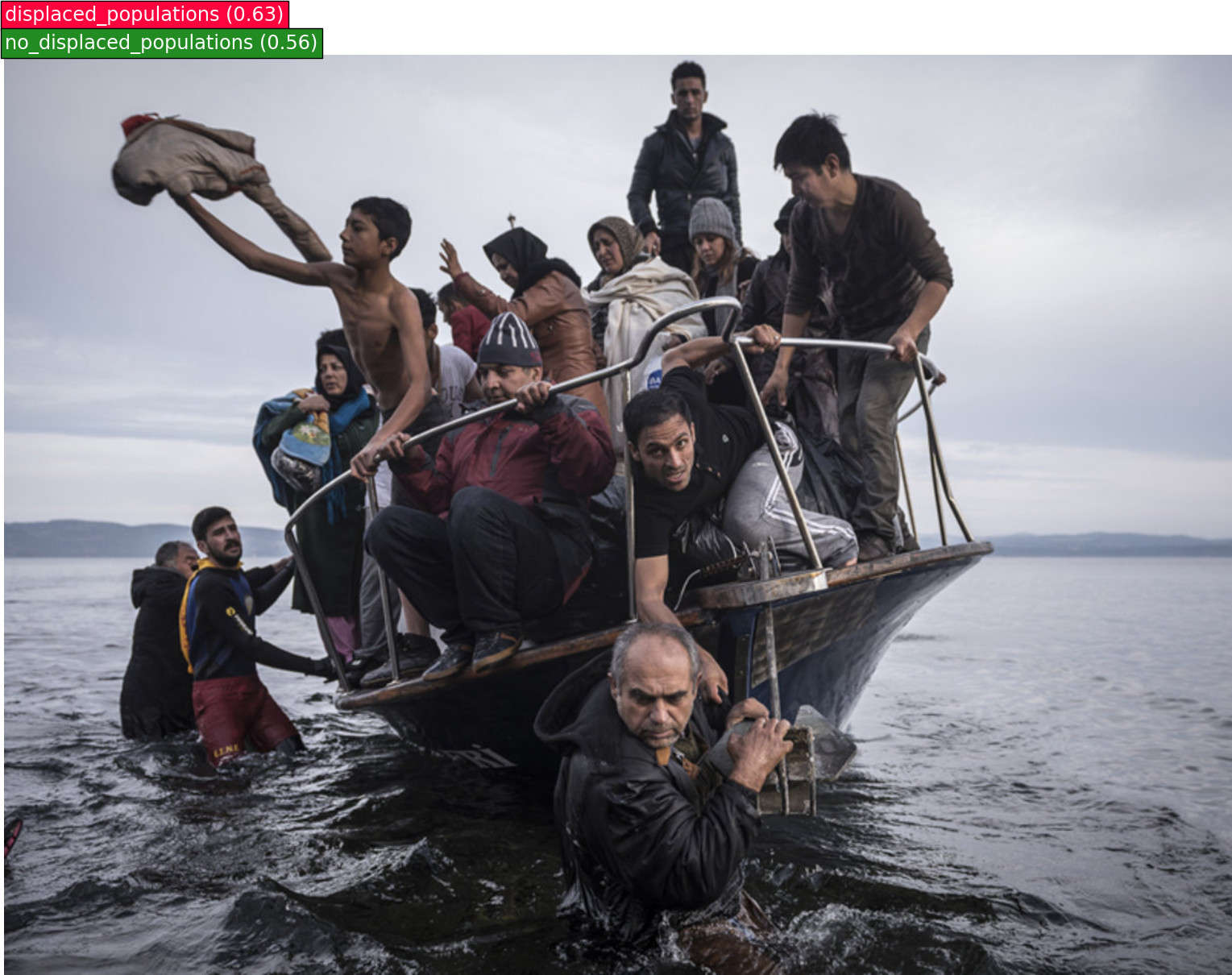} &
		\includegraphics[width=0.24\textwidth,height=0.15\textheight,keepaspectratio]{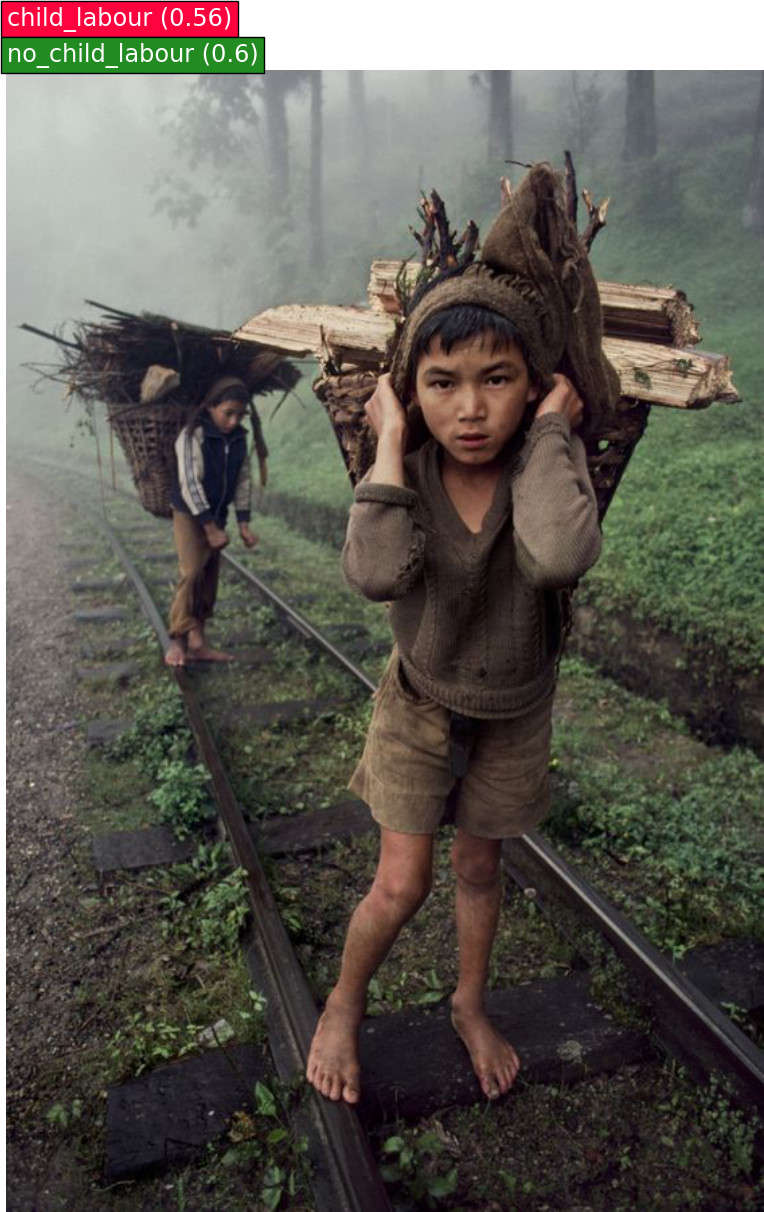} &
		\includegraphics[width=0.24\textwidth,height=0.42\textheight,keepaspectratio]{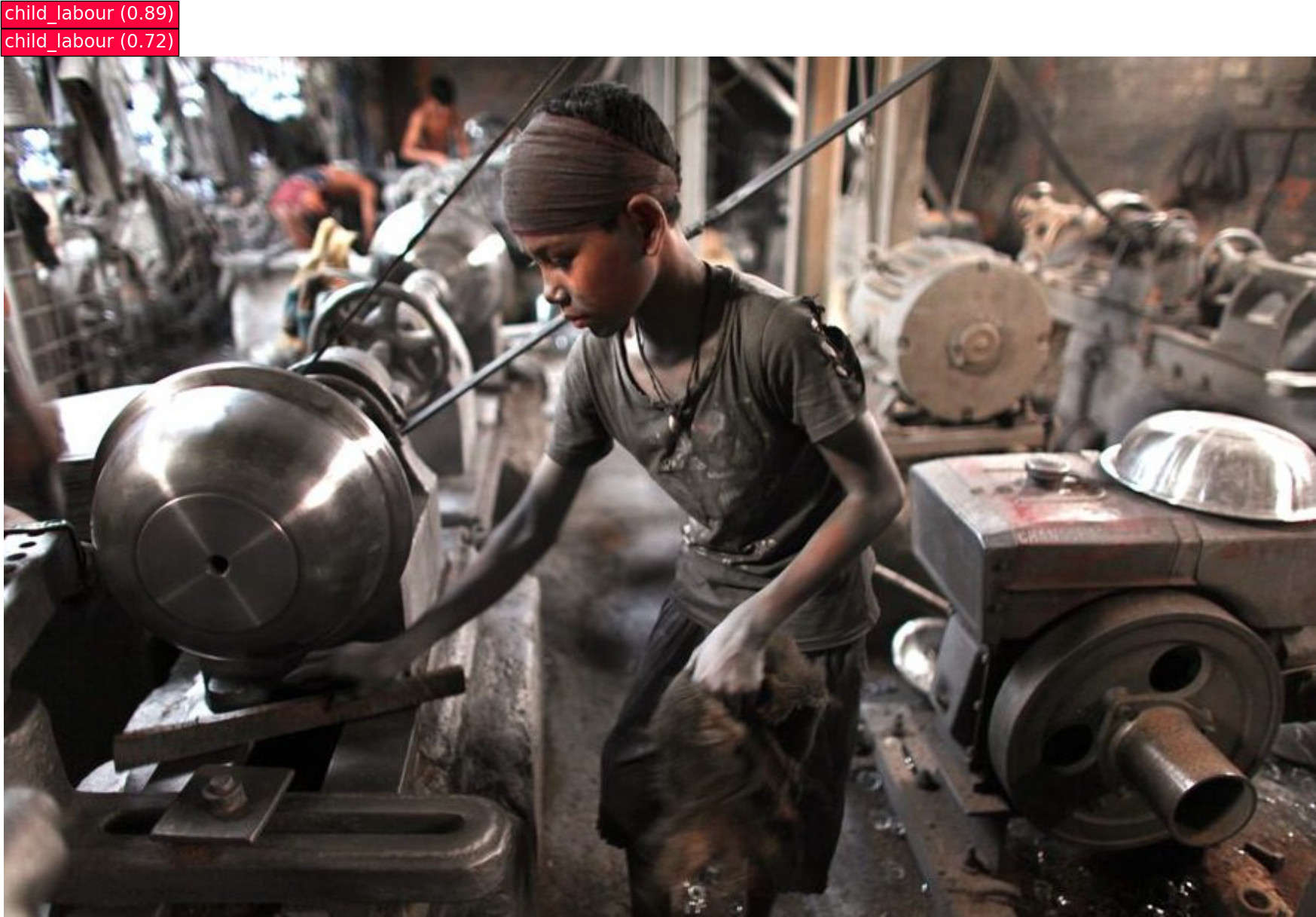} \\
		(a) & (b)  & (c) & (d) \\[-1pt]
		\includegraphics[width=0.24\textwidth,height=0.42\textheight,keepaspectratio]{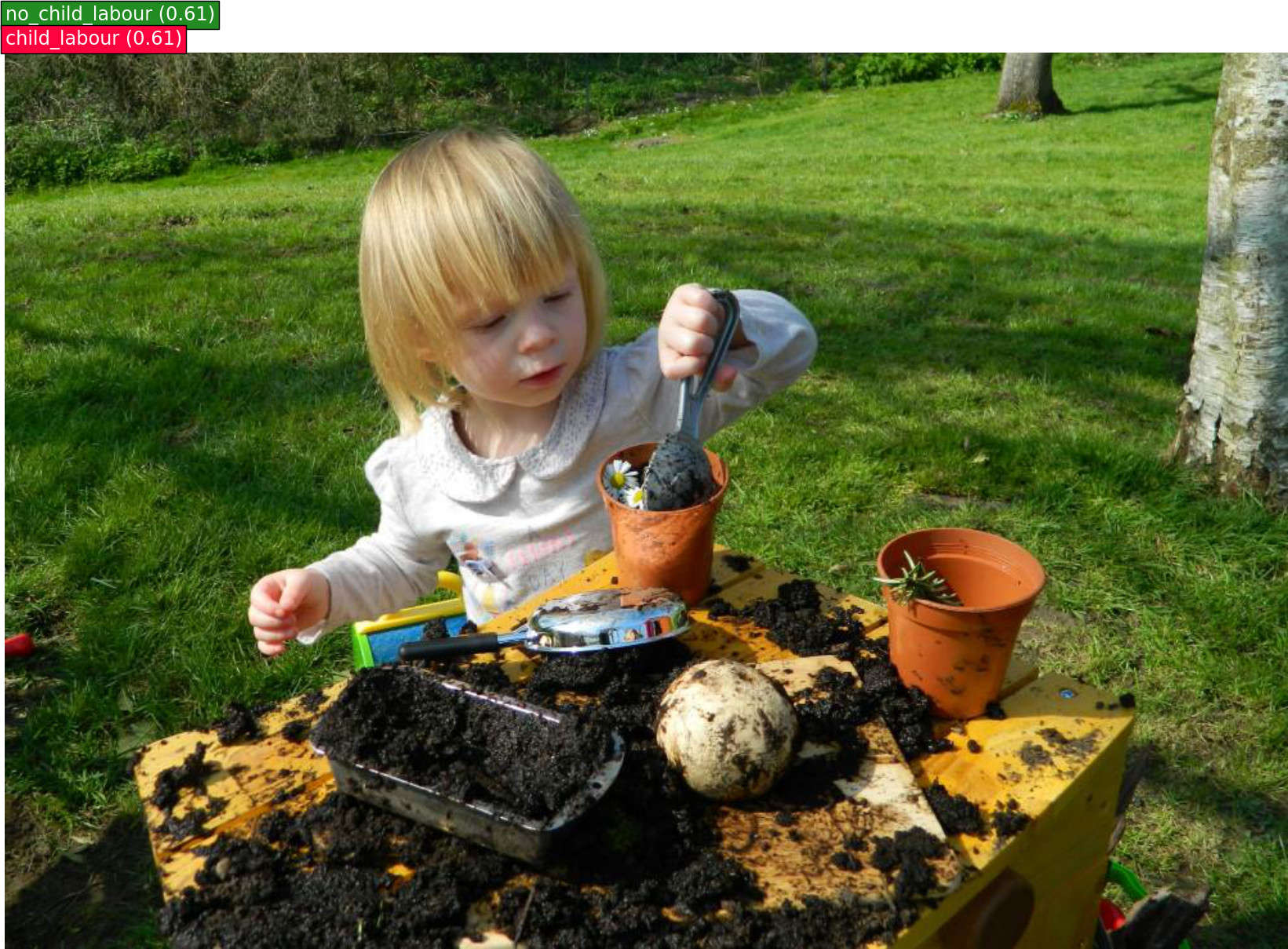} &   \includegraphics[width=0.24\textwidth,height=0.42\textheight,keepaspectratio]{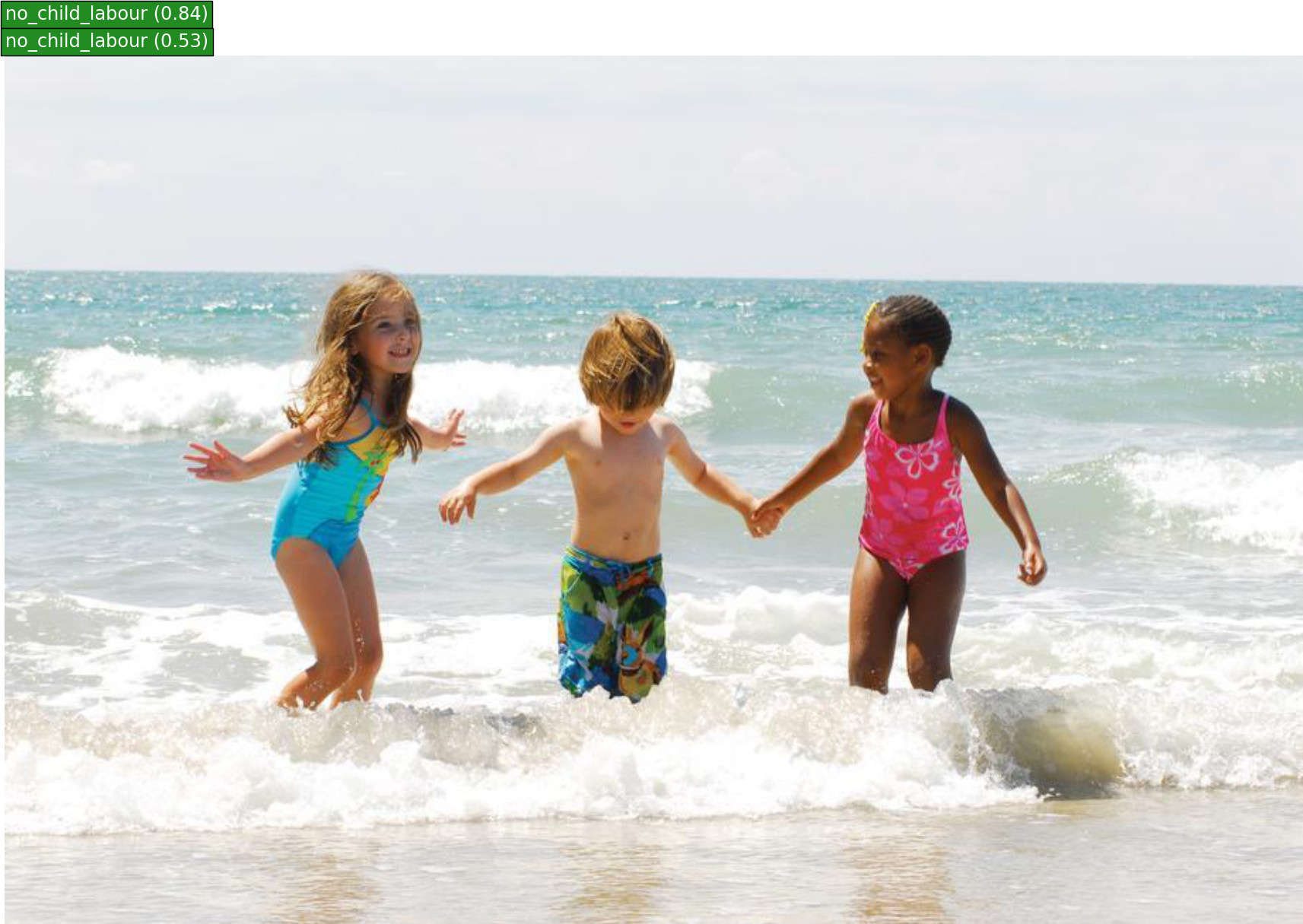} &
		\includegraphics[width=0.24\textwidth,height=0.42\textheight,keepaspectratio]{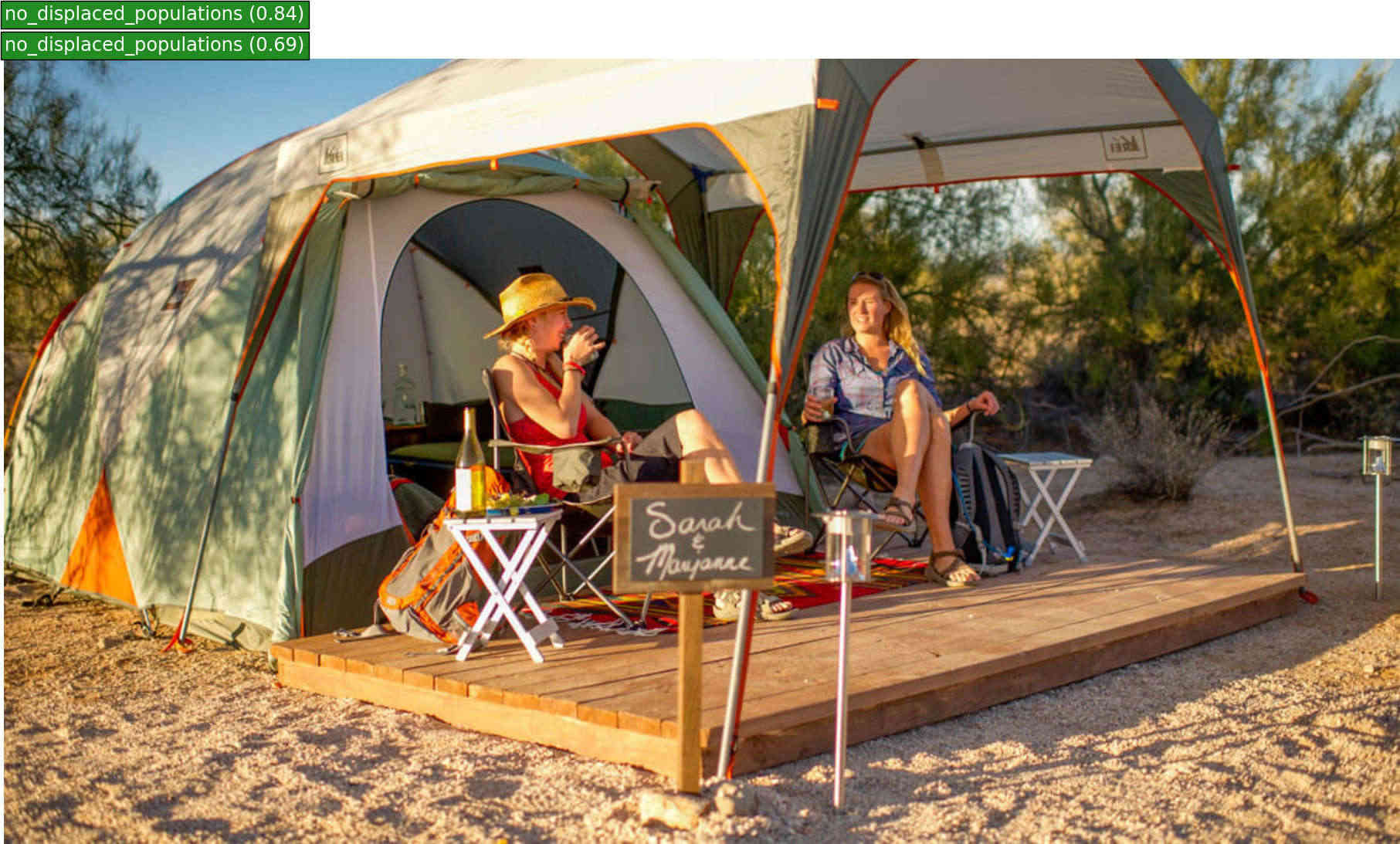} &
		\includegraphics[width=0.24\textwidth,height=0.42\textheight,keepaspectratio]{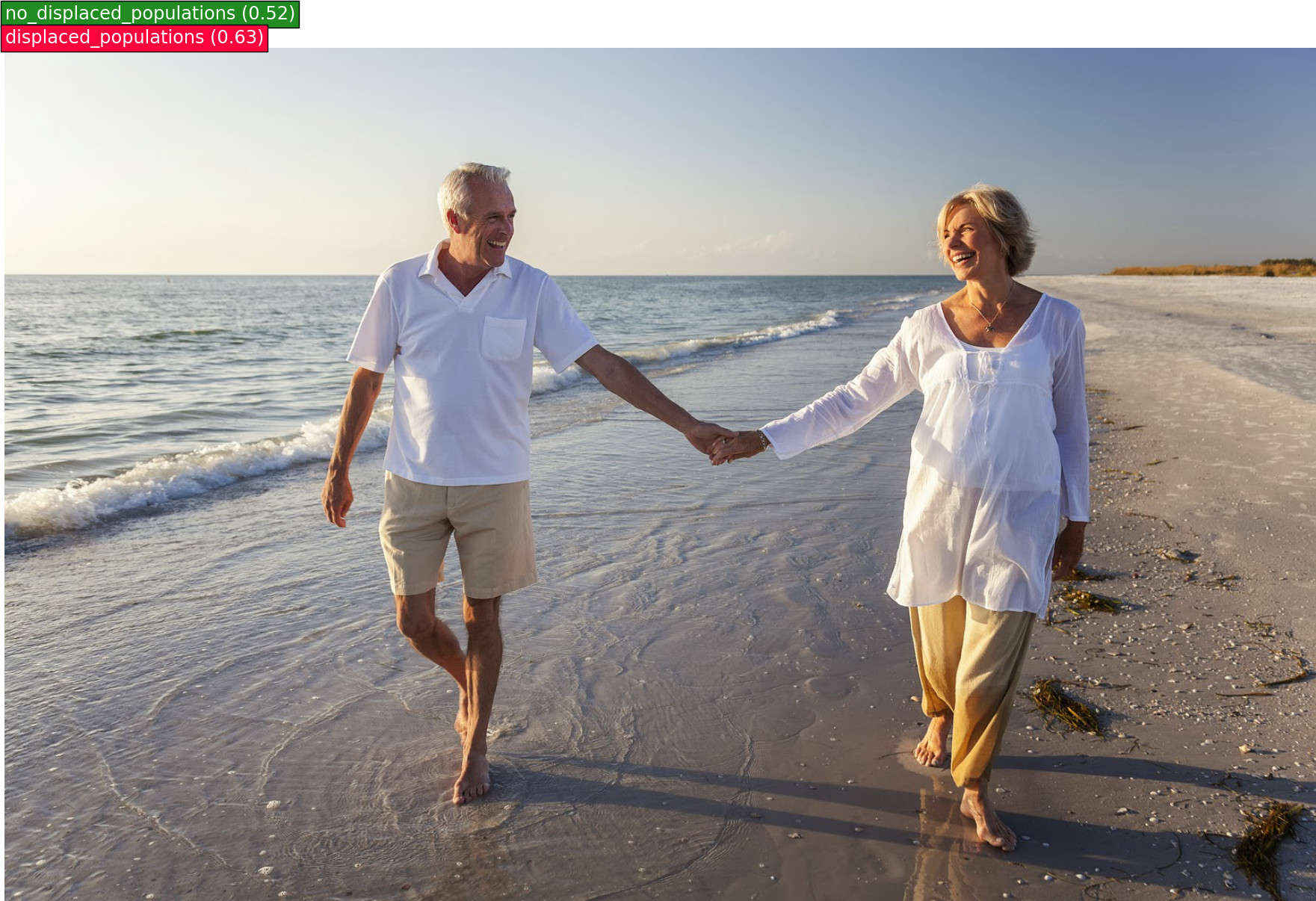} \\
		(e) & (f)  & (g)  & (h)  \\[1pt]
	\end{tabular}
	\caption{Human rights abuses detected by our method. Each image shows two predictions alongside their probabilities. Top prediction is given by GET-AID, while the bottom prediction is given by the respective vanilla CNN. Green colour implies that no human right abuse is detected, while red colour signifies that a human right abuse is detected. In some instances such as (a), (b), (c), (e) and (h), GET-AID overturns the initial-false prediction of the vanilla CNN, where in other instances such as (d), (f) and (g), GET-AID strengthens the initial-true prediction, resulting in higher coverage.}
	\label{Fig. 5}
\end{figure*}

\subsection{Baseline}
\label{subsec:baselines}
To enable a fair comparison between vanilla CNNs and GET-AID, we use the same backbone combinations for all modules described in Fig. \ref{Fig. 2}. We report comparisons in both \textit{accuracy} and \textit{coverage} metrics for fine-tuning up to three convolutional layers. The per-network results for \textit{child labour} and \textit{displaced populations} are shown in Table \ref{tab2} and Table \ref{tab3} respectively. The implementation of vanilla CNNs is solid: it has up to 65\% accuracy on the child labour classification and up to 80\% accuracy on the displaced populations classification, 29.82 and 44.82 points higher \footnote{Only cases with a single or two layers fine-tuned were considered for calculating these numbers in order to be consistent with the implementation of \cite{kalliatakis2018exploring}} than the 35.18 reported in the \textit{single-label}, \textit{multiclass classification} problem of \cite{kalliatakis2018exploring}. Regarding coverage, vanilla CNNs achieve up to 64\% for child labour classification and up to 67\% child displaced populations classification, which is in par with the 64\% maximum coverage reported in 
\cite{kalliatakis2018exploring}. We believe that this accuracy gap is mainly due to the fact that \cite{kalliatakis2018exploring} deals with a multiclass classification problem, whereas in this work we classify inputs into two mutually exclusive classes. This comparison showed that it is possible to trade coverage with accuracy in the context of human rights image analysis. One can always obtain high accuracy by refusing to process a number of examples, but this reduces the coverage of the system. Nevertheless, vanilla CNNs provide a strong baseline to which we will compare our method.

Our method, GET-AID, has a mean coverage of 64.75\% for child labour and 46.11\% for displaced populations on the HRA\textemdash Binary $test$ set. This is an absolute gain of 12.42 and 16.78 points over the strong baselines of 52.33\% and 29.33\% respectively. This is a relative improvement of 23.73\% and 57.21\% respectively. In relation to accuracy, GET-AID has a mean accuracy of 54\% for child labour and 61.25\% for displaced populations, which is an absolute drop of 4 and 5.5 points over the strong baselines of 58\% and 66.75\% accordingly. This indicates a relative loss of only 6.89\% and 8.23\% respectively. We believe that this negligible drop in accuracy is mainly due to the fact that the HRA\textemdash Binary $test$ set is not solely made up of images with people in their context, it also contains images of generic objects and scenes, where only the sole classifier's prediction is taken into account.

\begin{table}[t!]
	\caption{\label{tab2}Detailed results on HRA\textemdash Binary for the \textit{child labour} scenario. We show the main baseline and GET-AID for various network backbones. We bold the leading entries on coverage.}
	\centering
	\resizebox{\columnwidth}{!}{%
		\begin{tabular}{c|c|cc|cc}
			\multirow{2}{*}{\begin{tabular}[c]{@{}c@{}}backbone \\ network\end{tabular}} & \multirow{2}{*}{\begin{tabular}[c]{@{}c@{}}layers \\ fine-tuned\end{tabular}} & \multicolumn{2}{c|}{vanilla CNN} & \multicolumn{2}{c|}{\textbf{GET-AID}} \\ \cline{3-6} 
			&                                                                               & Top-1 acc.       & Coverage      & Top-1 acc.     & \textbf{Coverage}     \\ \hline
			VGG16                                                                        & \multirow{4}{*}{1}                                                            & 62\%             & 73\%            & 56\%           & \textbf{78\%}           \\
			VGG19                                                                        &                                                                               & 65\%             & 30\%            & 57\%           & \textbf{55\%}           \\
			ResNet50                                                                     &                                                                               & 51\%             & 0\%             & 50\%            & \textbf{24\%}           \\
			Places365                                                                    &                                                                               & 59\%             & 71\%            & 54\%           & \textbf{81\%}           \\ \hline
			VGG16                                                                        & \multirow{4}{*}{2}                                                            & 61\%             & 77\%            & 59\%           & \textbf{78\%}           \\
			VGG19                                                                        &                                                                               & 61\%             & 64\%            & 59\%           & \textbf{76\%}           \\
			ResNet50                                                                     &                                                                               & 52\%            & 0\%             & 49\%           & \textbf{33\%}           \\
			Places365                                                                    &                                                                               & 54\%             & 44\%            & 52\%           & \textbf{65\%}           \\ \hline
			VGG16                                                                        & \multirow{4}{*}{3}                                                            & 56\%             & 83\%            & 56\%           & \textbf{84\%}           \\
			VGG19                                                                        &                                                                               & 55\%             & \textbf{87\%}            & 55\%           & 82\%                    \\
			ResNet50                                                                     &                                                                               & 50\%              & \textbf{99\%}            & 48\%           & 91\%                    \\
			Places365                                                                    &                                                                               & 67\%             & 0\%             & 53\%           & \textbf{30\%}           \\ \hline
			\textbf{mean}                                                                        & -                                                            & 58\%             & 52.33\%            & 54\%           & \textbf{64.75\%}           \\ \hline
		\end{tabular}
	}
\end{table}

\begin{table}[t!]
	\caption{\label{tab3}Detailed results on HRA\textemdash Binary for the \textit{displaced populations} scenario. We show the main baseline and GET-AID for various network backbones. We bold the leading entries on coverage.}
	\centering
	\resizebox{\columnwidth}{!}{%
		\begin{tabular}{c|c|cc|cc}
			\multirow{2}{*}{\begin{tabular}[c]{@{}c@{}}backbone \\ network\end{tabular}} & \multirow{2}{*}{\begin{tabular}[c]{@{}c@{}}layers \\ fine-tuned\end{tabular}} & \multicolumn{2}{c|}{vanilla CNN} & \multicolumn{2}{c|}{\textbf{GET-AID}} \\ \cline{3-6} 
			&                                                                               & Top-1 acc.       & Coverage      & Top-1 acc.     & \textbf{Coverage}     \\ \hline
			VGG16                                                                        & \multirow{4}{*}{1}                                                            & 58\%             & 0\%           & 56\%           & \textbf{24.4\%}       \\
			VGG19                                                                        &                                                                               & 69\%             & 3\%           & 59\%           & \textbf{33\%}         \\
			ResNet50                                                                     &                                                                               & 60\%             & 0\%           & 53\%           & \textbf{29\%}         \\
			Places365                                                                    &                                                                               & 64\%             & 3\%           & 54\%           & \textbf{32\%}         \\ \hline
			VGG16                                                                        & \multirow{4}{*}{2}                                                            & 63\%             & 43\%          & 60\%           & \textbf{58\%}         \\
			VGG19                                                                        &                                                                               & 77\%             & 54\%          & 70\%           & \textbf{59\%}         \\
			ResNet50                                                                     &                                                                               & 42\%             & 1\%           & 44\%           & \textbf{33\%}         \\
			Places365                                                                    &                                                                               & 80\%             & 49\%          & 73\%           & \textbf{58\%}         \\ \hline
			VGG16                                                                        & \multirow{4}{*}{3}                                                            & 72\%             & 69\%          & 67\%           & \textbf{71\%}         \\
			VGG19                                                                        &                                                                               & 82\%             & 64\%          & 77\%           & \textbf{68\%}         \\
			ResNet50                                                                     &                                                                               & 53\%             & 0\%           & 51\%           & \textbf{22\%}         \\
			Places365                                                                    &                                                                               & 81\%             & 66\%          & 71\%           & \textbf{66\%}         \\ \hline
			\textbf{mean}                                                                        & -                                                            & 66.75\%             & 29.37\%            & 61.25\%           & \textbf{46.11\%}           \\ \hline
		\end{tabular}
	}
\end{table}

\subsection{Qualitative results}
\label{subsec:qualitative_results}

\begin{figure*}[t!]
	\centering
	\begin{tabular}{cc}
		\includegraphics[width=0.35\textwidth,height=0.42\textheight,keepaspectratio]{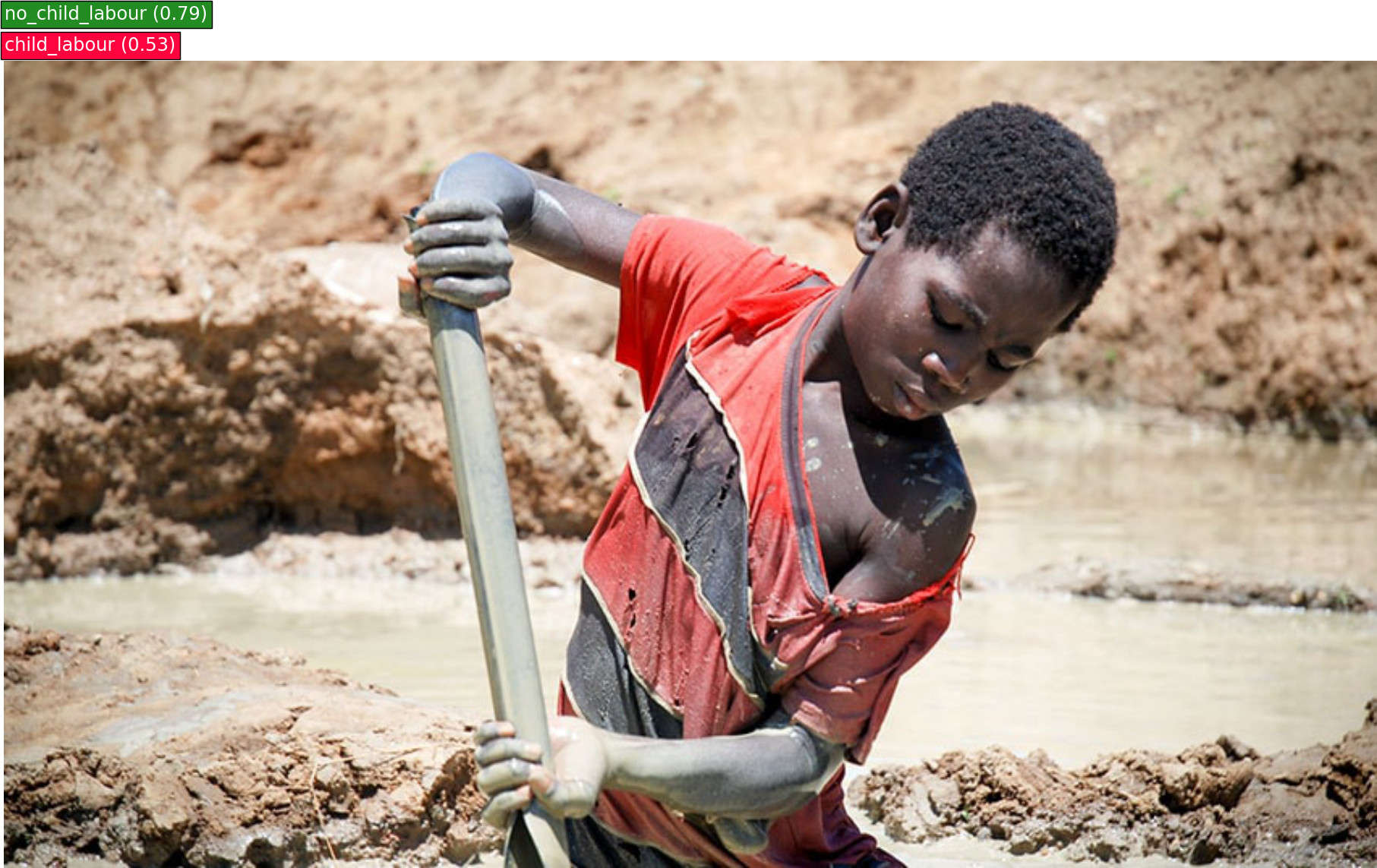} &   \includegraphics[width=0.3\textwidth,height=0.35\textheight,keepaspectratio]{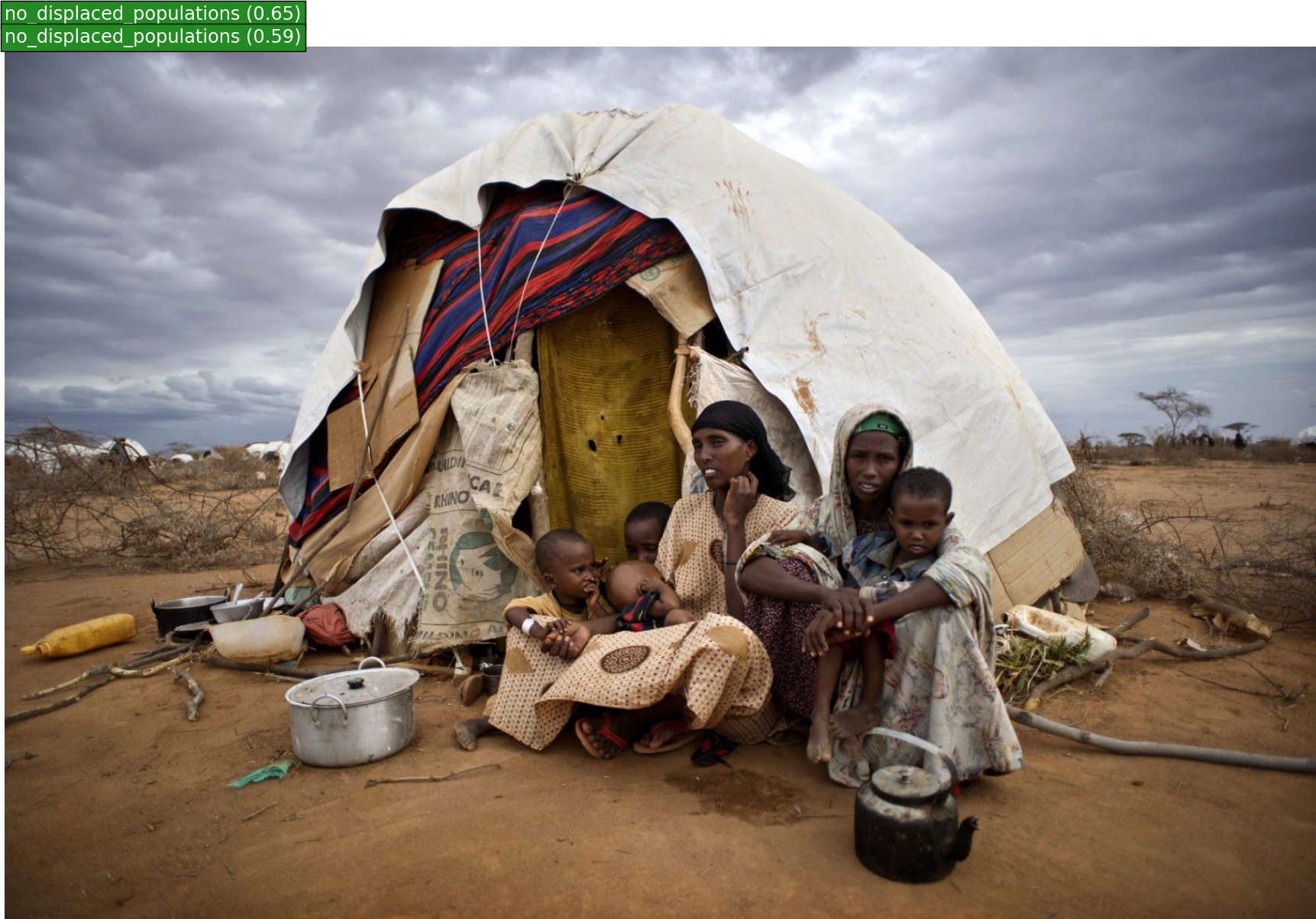} \\
		(a) & (b)  \\[6pt]
	\end{tabular}
	\caption{False positive detections of our method.}
	\label{Fig. 6}
\end{figure*}

We show our human rights abuse detection results in Fig. \ref{Fig. 5}. Each subplot illustrates two predictions alongside their probability scores. The top of the two predictions is given by GET-AID, while the bottom one is given by the respective vanilla CNN sole classifier. Our method can successfully classify human right abuses by overturning the initial-false prediction of the vanilla CNN in Fig. \ref{Fig. 5} (a), (b), (c), (e) and (h). Moreover, GET-AID can strengthen the initial-true prediction of the sole classifier as shown in Fig. \ref{Fig. 5} (d), (f) and (g) respectively.

\subsection{Failure cases}
\label{subsec:failure_cases}
Fig. \ref{Fig. 6} shows some failure detections. Our method can be incorrect, because of false global emotional traits inferences. Some of them are caused by a failure of continuous dimensions emotion recognition, which is an interesting open problem for future research.

\section{Conclusion}
\label{sec:conclusion}
We have presented a human-centric approach for classifying two types of human rights abuses. This two-class human rights abuse labelling problem is not trivial, given the high-level image interpretation required. Understanding what a person is experiencing from his frame of reference is closely related with situations where human rights are being violated. Thus, the key to our computational framework are people's emotional traits, which resonate well with our own common sense in judging potential human right abuses. We introduce \textit{global emotional traits} of the image which are responsible for weighting the classifiers prediction during inference. We benchmark performance of our GET-AID model against sole CNN classifiers. Our experimental results showed that this is an effective strategy, which we believe has good potential beyond human rights abuses classification. We hope this paper will spark interest and subsequent
work along this line of research. All our code and data are publicly available.

\vskip 0pt plus -1fil
\begin{IEEEbiography}[{\includegraphics[width=1in,height=1.25in,clip,keepaspectratio]{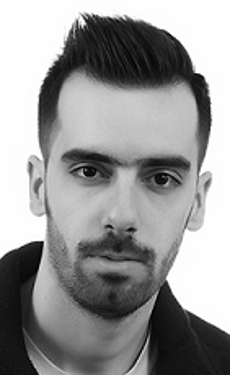}}]{\textbf{GRIGORIOS KALLIATAKIS}} received the B.Sc. and M.Sc. degrees in informatics engineering and informatics \& multimedia from Technological Education Institute (TEI) of Crete, Greece, in 2012 and 2015, respectively, and a M.Sc. degree in computer vision from University of Burgundy, France, in 2015. 
He has been pursuing the PhD in Computer Science from the University of Essex, UK, since 2015. He is currently working on visual recognition of human rights abuses. His research interests include computer vision, image interpretation and visual recognition.
\end{IEEEbiography}
\vskip 0pt plus -1fil

\begin{IEEEbiography}[{\includegraphics[width=1in,height=1.25in,clip,keepaspectratio]{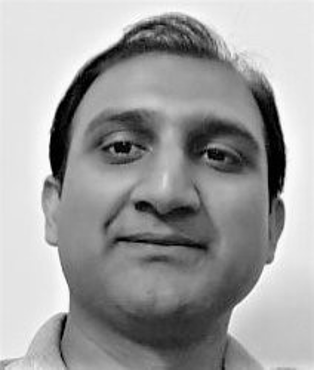}}]{\textbf{SHOAIB EHSAN}} received the B.Sc. degree in electrical engineering from the University of Engineering and Technology, Taxila, Pakistan, in 2003, and the Ph.D. degree in computing and electronic systems (with specialization in computer vision) from the University of Essex, Colchester, U.K., in 2012. He has an extensive industrial and academic experience in the areas of embedded systems, embedded software design, computer vision, and image processing. His current research interests are in intrusion detection for embedded systems, local feature detection and description techniques, and image feature matching and performance analysis of vision systems. He was a recipient of the University of Essex Post Graduate Research Scholarship and the Overseas Research Student Scholarship and also a recipient of the prestigious Sullivan Doctoral Thesis Prize awarded annually by the British Machine Vision Association.
\end{IEEEbiography}

\begin{IEEEbiography}[{\includegraphics[width=1in,height=1.25in,clip,keepaspectratio]{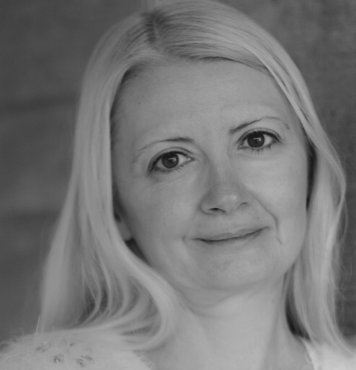}}]{\textbf{MARIA FASLI}} is currently a Professor of Computer Science (Artificial Intelligence) and the Director of the Institute for Analytics and Data Science (IADS) at the University of Essex. She obtained her BSc in Informatics from the Technological Education Institute of Thessaloniki in 1996, and her PhD in Computer Science from the University of Essex in 2000. She has held research and academic positions at the University of Essex since 1999 and became Professor in 2012. In 2009, she became the Head of the School of Computer Science and Electronic Engineering at Essex, a post which she held until the end of 2014. In August 2014, she was appointed in her current role as Director of IADS. In 2016, she was awarded the first UNESCO Chair in Analytics and Data Science. Her research interests lie in artificial intelligence techniques for analyzing and modeling complex systems and structured and unstructured data in various domains. Her research has been funded by National Research Councils in the UK, InnovateUK as well as businesses.
\end{IEEEbiography}
\vskip 0pt plus -1fil

\begin{IEEEbiography}[{\includegraphics[width=1in,height=1.55in,clip,keepaspectratio]{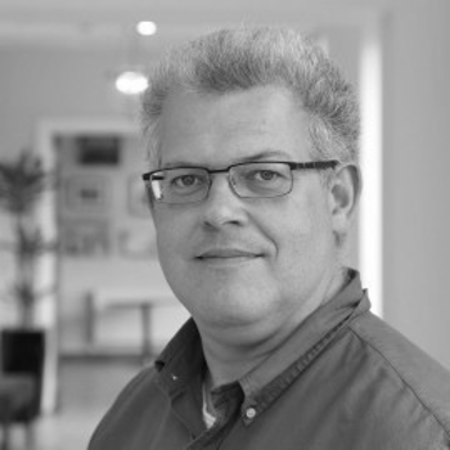}}]{\textbf{KLAUS D. MCDONALD-MAIER}}(S'91- SM'06) is currently the Head of the Embedded and Intelligent Systems Laboratory, University of Essex, Colchester, U.K. He is also Chief Scientist of UltraSoC Technologies Ltd., CEO of Metrarc Ltd., and a Visiting Professor with the University of Kent. His current research interests include embedded systems and system-on-chip design, security, development support and technology, parallel and energy efficient architectures, computer vision, data analytics, and the application of soft computing and image processing techniques for real world problems. He is a member of the VDE and a Fellow of the IET.
\end{IEEEbiography}

\EOD

\end{document}